\documentclass[sigplan,10pt]{acmart}
\renewcommand\footnotetextcopyrightpermission[1]{}
\AtBeginDocument{%
  }

\usepackage{subfigure}
\usepackage{color}
\usepackage{xcolor}
\usepackage{comment}
\usepackage{algorithm,algpseudocode}
\usepackage{rotating}
\usepackage{multirow}
\usepackage{tabularray}
\usepackage{enumitem}
\usepackage{url}
\usepackage{xspace}
\usepackage{subfigure}
\usepackage{amsmath}
\usepackage{tabularx}  
\usepackage{graphicx}
\usepackage{array}
\usepackage{balance}
\usepackage{newtxmath}

\usepackage{listings}

\usepackage{float}   
\usepackage{caption} 

\definecolor{codebg}{RGB}{248,248,248}
\definecolor{bittersweet}{rgb}{1.0, 0.44, 0.37}

\lstdefinestyle{pythonstyle}{
    language=Python,
    backgroundcolor=\color{codebg},
    basicstyle=\ttfamily\small,
    keywordstyle=\color{blue}\bfseries,
    stringstyle=\color{orange},
    commentstyle=\color{gray},
    showstringspaces=false,
    breaklines=true,
    frame=single,
    rulecolor=\color{black!20},
    tabsize=4,
    numbers=left,
    numberstyle=\tiny\color{gray},
    stepnumber=1,
    numbersep=10pt
}

\newcommand{\name}{\textsc{UFO}\textsuperscript{2}\xspace}
\newcommand{\hosta}{\textsc{HostAgent}\xspace}
\newcommand{\appa}{\textsc{AppAgent}\xspace}

\newcommand{\appas}{\textsc{AppAgent}s\xspace}

\newcommand{\eg}{\textit{e.g.},\xspace}
\newcommand{\ie}{\textit{i.e.},\xspace}
\newcommand{\etal}{\textit{et al.},\xspace}

\newcommand{\intern}{\authornote{This work was done during their internship at Microsoft.}}
\newcommand{\correspond}{\authornote{Corresponding author. Email: \texttt{chaoyun.zhang@microsoft.com}}}

\usepackage{etoolbox}      

\begin{document}

\title{\name: The Desktop AgentOS}


\settopmatter{authorsperrow=5}

\author{Chaoyun Zhang}
\correspond
\affiliation{%
  \institution{Microsoft}
  \country{}
}

\author{He Huang}
\affiliation{%
  \institution{Microsoft}
  \country{}
}

\author{Chiming Ni}
\intern
\affiliation{%
  \institution{ZJU-UIUC Institute}
  \country{}
}

\author{Jian Mu\footnotemark[2]}
\affiliation{%
  \institution{Nanjing University}
  \country{}
}

\author{Si Qin}
\affiliation{%
  \institution{Microsoft}
  \country{}
}

\author{Shilin He}
\affiliation{%
  \institution{Microsoft}
  \country{}
}

\author{Lu Wang}
\affiliation{%
  \institution{Microsoft}
  \country{}
}

\author{Fangkai Yang}
\affiliation{%
  \institution{Microsoft}
  \country{}
}

\author{Pu Zhao}
\affiliation{%
  \institution{Microsoft}
  \country{}
}

\author{Bo Qiao}
\affiliation{%
  \institution{Microsoft}
  \country{}
}

\author{Chao Du}
\affiliation{%
  \institution{Microsoft}
  \country{}
}

\author{Liqun Li}
\affiliation{%
  \institution{Microsoft}
  \country{}
}

\author{Yu Kang}
\affiliation{%
  \institution{Microsoft}
  \country{}
}

\author{Zhao Jiang}
\affiliation{%
  \institution{Microsoft}
  \country{}
}

\author{Suzhen Zheng}
\affiliation{%
  \institution{Microsoft}
  \country{}
}

\author{Rujia Wang}
\affiliation{%
  \institution{Microsoft}
  \country{}
}

\author{Jiaxu Qian\footnotemark[2]}
\affiliation{%
  \institution{Peking University}
  \country{}
}

\author{Minghua Ma}
\affiliation{%
  \institution{Microsoft}
  \country{}
}

\author{Jian-Guang Lou}
\affiliation{%
  \institution{Microsoft}
  \country{}
}

\author{Qingwei Lin}
\affiliation{%
  \institution{Microsoft}
  \country{}
}

\author{Saravan Rajmohan}
\affiliation{%
  \institution{Microsoft}
  \country{}
}

\author{Dongmei Zhang}
\affiliation{%
  \institution{Microsoft}
  \country{}
}

\renewcommand{\shortauthors}{Chaoyun Zhang \textit{et al.}}   

\makeatletter
\def\@mkbibcitation{\bgroup
  \let\@vspace\@vspace@orig
  \let\@vspacer\@vspacer@orig
  \def\@pages@word{\ifnum\getrefnumber{TotPages}=1\relax page\else pages\fi}%
  \def\footnotemark{}%
  \def\\{\unskip{} \ignorespaces}%
  \def\footnote{\ClassError{\@classname}{Please do not use footnotes
      inside a \string\title{} or \string\author{} command! Use
      \string\titlenote{} or \string\authornote{} instead!}}%
  \def\@article@string{\ifx\@acmArticle\@empty{\ }\else,
    Article~\@acmArticle\ \fi}%
  \par\medskip\small\noindent{\bfseries ACM Reference Format:}\par\nobreak
  \noindent\bgroup
    \def\\{\unskip{}, \ignorespaces}\shortauthors\egroup. \@acmYear. \@title
  \ifx\@subtitle\@empty. \else: \@subtitle. \fi
  \if@ACM@nonacm\else
    \if@ACM@journal@bibstrip
       \textit{\@journalNameShort}
       \@acmVolume, \@acmNumber \@article@string (\@acmPubDate),
       \ref{TotPages}~\@pages@word.
    \else
       In \textit{\@acmBooktitle}%
       \ifx\@acmEditors\@empty\textit{.}\else
         \andify\@acmEditors\textit{, }\@acmEditors~\@editorsAbbrev.%
       \fi\
       ACM, New York, NY, USA%
         \@article@string\unskip, \ref{TotPages}~\@pages@word.
    \fi
  \fi
  \ifx\@acmDOI\@empty\else\@formatdoi{\@acmDOI}\fi
\par\egroup}
\makeatother


\begin{abstract}
    Recent \emph{Computer-Using Agents} (CUAs), powered by multimodal large language models (LLMs), offer a promising direction for automating complex desktop workflows through natural language. However, most existing CUAs remain conceptual prototypes, hindered by shallow OS integration, fragile screenshot-based interaction, and disruptive execution.

    We present \textbf{\name}, a multiagent \emph{AgentOS} for Windows desktops that elevates CUAs into practical, system-level automation. \name features a centralized \hosta for task decomposition and coordination, alongside a collection of application-specialized \appas equipped with native APIs, domain-specific knowledge, and a unified GUI--API action layer. This architecture enables robust task execution while preserving modularity and extensibility. A hybrid control detection pipeline fuses Windows UI Automation (UIA) with vision-based parsing to support diverse interface styles. Runtime efficiency is further enhanced through speculative multi-action planning, reducing per-step LLM overhead. Finally, a \emph{Picture-in-Picture} (PiP) interface enables automation within an isolated virtual desktop, allowing agents and users to operate concurrently without interference.

    We evaluate \name across over 20 real-world Windows applications, demonstrating substantial improvements in robustness and execution accuracy over prior CUAs. Our results show that deep OS integration unlocks a scalable path toward reliable, user-aligned desktop automation.

    The source code of \name is publicly available at {\color{bittersweet}\url{https://github.com/microsoft/UFO/}}, with comprehensive documentation provided at {\color{bittersweet}\url{https://microsoft.github.io/UFO/}}.
\end{abstract}
\keywords{Computer Using Agent, Large Language Model, Desktop Automation, Windows System}

\maketitle
\settopmatter{printfolios=true}

\begin{figure}[t]
    \centering
    \includegraphics[width=\columnwidth]{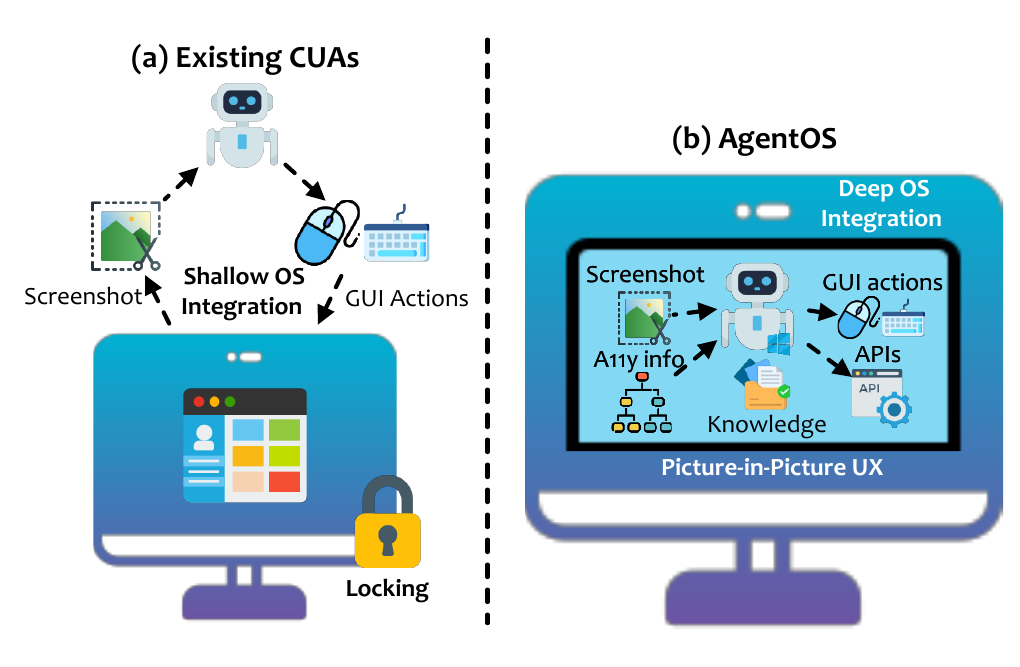}
    \vspace{-2.5em}
    \caption{A comparison of (a) existing CUAs and (b) desktop AgentOS \name.}
    \label{fig:comparison}
\end{figure}

\section{Introduction}

Automation of desktop applications has long been central to improving workforce productivity. Commercial Robotic Process Automation (RPA) platforms such as UiPath~\cite{UiPath}, Automation Anywhere~\cite{AutomationAnywhere}, and Microsoft Power Automate~\cite{PowerAutomate} exemplify this trend, using predefined scripts to replicate repetitive user interactions through the graphical user interface (GUI) \cite{hofmann2020robotic, madakam2019future}. However, these script-based approaches often prove fragile in dynamic, continuously evolving environments \cite{pramod2022robotic}. Minor interface changes can break the underlying automation scripts, requiring manual updates and extensive maintenance effort. As software ecosystems grow increasingly complex and heterogeneous, the brittleness of script-based automation severely limits scalability, adaptability, and practicality.

Recently, \emph{Computer-Using Agents} (CUAs) \cite{zhang2024large} have emerged as a promising alternative. These systems leverage advanced multimodal large language models (LLMs) \cite{zhao2023survey, zhang2024mm} to interpret diverse user instructions, perceive GUI interfaces, and generate adaptive actions (\eg mouse clicks, keyboard inputs) without fixed scripting~\cite{zhang2024large}. Early prototypes such as UFO~\cite{zhang2024ufo}, Anthropic Claude~\cite{anthropic2024}, and OpenAI Operator~\cite{cua2025} demonstrate that such LLM-driven agents can robustly automate tasks too complex or ambiguous for conventional RPA pipelines. Yet, despite these advances, current CUA implementations remain largely \emph{conceptual}: they mainly optimize visual grounding or language reasoning~\cite{wang2024large, qin2025ui, cua2025, zheng2025vem, niu2024screenagent}, but give little attention to \emph{system-level} integration with desktop operating systems (OSs) and application internals (Figure~\ref{fig:comparison}~(a)).

Reliance on raw GUI screenshots and simulated input events has several drawbacks. First, visual-only inputs can be noisy and redundant, increasing cognitive overhead for LLMs and reducing execution efficiency~\cite{zhang2025apiagentsvsgui}. Second, existing CUAs rarely leverage the OS's native accessibility interfaces, application-level APIs, or detailed process states---a missed opportunity to significantly enhance decision accuracy, reduce latency, and enable more reliable execution. Finally, simulating mouse and keyboard events on the primary user desktop locks users out during automation, imposing a poor user experience (UX). These limitations must be resolved before CUAs can mature from intriguing prototypes into robust, scalable solutions for real-world desktop automation, and motivates our central research question:
\begin{quote} 
    \textit{How can we build a robust, deeply integrated system for desktop automation that flexibly adapts to evolving interfaces, reliably orchestrates diverse applications, and minimizes disruption to user workflows?}
\end{quote}

In response, we present \textbf{\name}, a new \emph{AgentOS} for Windows that reimagines desktop automation as a first-class operating system abstraction. Unlike prior CUAs that treat automation as a layer atop screenshots and simulated input events, \name is architected as a deeply integrated, multiagent execution environment—embedding OS capabilities, application-specific introspection, and domain-aware planning into the core automation loop, as illustrated in Figure~\ref{fig:comparison}(b).

At its foundation, \name provides a modular, system-level substrate for natural language-driven automation. A centralized coordinator, the \hosta, interprets user instructions, decomposes them into semantically meaningful subtasks, and dynamically dispatches execution to specialized \appas—expert modules tailored for specific Windows applications. Each \appa is equipped with an extensible toolbox of application-specific APIs, a hybrid GUI–API action interface, and integrated knowledge about the application's capabilities and semantics. This architecture enables robust orchestration across multiple concurrent applications, supporting workflows that span Excel, Outlook, Edge, and beyond.

To enable reliable execution across the full spectrum of application UIs, \name introduces a hybrid control detection pipeline, combining Windows UI Automation (UIA) APIs with advanced visual grounding models~\cite{lu2024omniparser}. This allows agents to introspect and act on both standard and custom UI components, bridging the gap between structured accessibility trees and pixel-level perception. Moreover, \name continuously incorporates external documentation, patch notes, and past execution traces into a unified vectorized memory layer, enabling each \appa to incrementally refine its behavior without retraining.

At the interaction layer, \name exposes a unified GUI–API execution model, where agents seamlessly combine traditional GUI actions (\eg clicks, keystrokes) with native Windows or application-specific APIs. This hybrid approach improves execution efficiency, reduces brittleness to UI layout changes, and enables more expressive, higher-level operations. To further minimize the latency associated with LLM-based action planning, \name incorporates a speculative multi-action execution engine that proactively infers and validates action sequences using lightweight control-state checks at a single inference step—substantially reducing inference overhead without compromising correctness.

Finally, to ensure a practical and non-intrusive user experience, \name introduces a novel Picture-in-Picture (PiP) interface: a secure, nested desktop environment where agents can execute independently of the user's main session. Built atop Windows' native remote desktop loopback infrastructure, PiP enables seamless, side-by-side user–agent multitasking without disruption on the user's main desktop, addressing one of the most persistent UX limitations of existing CUAs.

Together, these design principles position \name not just as a smarter agent, but as a new OS-level abstraction for automation—transforming desktop workflows into programmable, composable, and robust entities. In summary, this paper makes the following contributions:
\begin{itemize} 
    \item \textbf{Deep OS Integration:} We design and implement \name, a multiagent AgentOS that deeply embeds within the Windows OS, orchestrating desktop applications through introspection, API access, and fine-grained execution control. 
    \item \textbf{Unified GUI--API Action Layer:} We propose a hybrid action interface that bridges traditional GUI interactions with application-native API calls, enabling flexible, efficient, and robust automation.
    \item \textbf{Hybrid Control Detection:} We introduce a fusion pipeline combining UIA metadata with vision-based detection to achieve reliable control grounding even in non-standard interfaces.
    \item \textbf{Continuous Knowledge Integration:} We build a retrieval-augmented memory that integrates documentation and historical execution logs, allowing agents to improve autonomously over time without retraining.
    \item \textbf{Speculative Multi-Action Execution:} We reduce LLM invocation overhead by predicting and validating action sequences ahead of time using UI state signals.
    \item \textbf{Non-Disruptive UX:} We develop a nested virtual desktop environment that allows automation to proceed in parallel with user activity, avoiding interference and improving adoptability.
    \item \textbf{Comprehensive Evaluation:} We evaluate \name across 20+ real-world Windows applications, showing consistent improvements in success rate, execution efficiency, and usability over state-of-the-art CUAs like Operator.
\end{itemize}
Overall, \name advances the vision of OS-native automation by shifting the paradigm from GUI scripting to structured, programmable application control. Even when paired with general-purpose models like GPT-4o, \name outperforms dedicated CUAs by over 10\%, highlighting the transformative impact of system-level integration and architectural design.

\section{Background}\label{sec:background}

\subsection{The Fragility of Traditional Desktop Automation}

For decades, desktop automation has relied on brittle techniques to replicate human interactions with GUI-based applications. Commercial RPA (Robotic Process Automation) tools—such as UiPath~\cite{UiPath}, Automation Anywhere~\cite{AutomationAnywhere}, and Microsoft Power Automate~\cite{PowerAutomate}—operate by recording and replaying mouse movements, keystrokes, or rule-based scripts. These systems rely heavily on surface-level GUI cues (\eg pixel regions, window titles), offering little introspection into application state.

While widely deployed in enterprise settings, traditional RPA systems exhibit poor robustness and scalability~\cite{siderska2023towards}. Even minor UI updates—such as reordering buttons or relabeling menus—can silently break automation scripts. Maintaining correctness requires frequent human intervention. Furthermore, these tools lack semantic understanding of application workflows and cannot reason about or adapt to novel tasks. As a result, RPA tools remain constrained to narrow, repetitive workflows in stable environments, far from general-purpose automation.

\subsection{Rise of Computer-Using Agents}

Recent advances in large language models (LLMs) and multimodal perception have enabled a new class of automation systems, referred to as \emph{Computer-Using Agents} (CUAs)~\cite{zhang2024large, zhao2023survey, zhang2024mm}. CUAs aim to generalize across applications and tasks by leveraging LLMs to interpret user instructions, perceive GUI layouts, and synthesize actions such as clicks and keystrokes. Early CUAs like UFO~\cite{zhang2024ufo} demonstrated that multimodal models (\eg GPT-4V~\cite{yang2023dawn}) could map natural language requests to sequences of GUI actions with no hand-crafted scripts. More recent industry prototypes, including Claude-3.5 (Computer Use)~\cite{anthropic2024} and OpenAI Operator~\cite{cua2025}, have pushed the envelope further, performing realistic desktop workflows across multiple applications.

These CUAs represent a promising evolution from static RPA scripts to adaptive, general-purpose automation. However, despite their sophistication, current CUAs largely remain research prototypes, constrained by architectural and systems-level limitations that impede real-world deployment.

\subsection{Systems Challenges in CUAs}

Current CUAs fall short in three fundamental ways, which we argue stem from missing operating system abstractions:

\paragraph{(1) Lack of OS-Level Integration.}
Most CUAs interact with the system through screenshots and low-level input emulation (mouse and keyboard events). They ignore rich system interfaces such as accessibility APIs, application process state, and native inter-process communication mechanisms (\eg shell commands, COM interfaces~\cite{gray1998modern}). This superficial interaction model limits reliability and efficiency—every action must be inferred from pixels rather than structured state.

\paragraph{(2) Absence of Application Introspection.}
CUAs typically operate as generalists with limited awareness of application-specific capabilities. They treat all interfaces uniformly, lacking the ability to leverage built-in APIs or vendor documentation. As a result, they cannot reason over high-level concepts unless such flows are explicitly embedded in the model. This rigidity limits their generalization and makes maintenance expensive.

\paragraph{(3) Disruptive and Unsafe Execution Model.}
Most CUAs drive automation directly on the user's desktop session, hijacking the real mouse and keyboard. This design prevents users from interacting with their system during execution, introduces interference risk, and violates isolation principles fundamental to safe system design. Long-running tasks—especially those involving multiple LLM queries—can monopolize the session for minutes at a time.

\subsection{Missing Abstraction: OS Support for Automation}

Despite growing demand for intelligent, language-driven automation, existing operating systems offer no first-class abstraction for exposing GUI application control to external agents. In contrast to system calls, files, or sockets, GUI workflows remain opaque and non-programmable. As a result, both RPA and CUA systems are forced to operate as ad-hoc layers atop the GUI, with no unified substrate for execution, coordination, or introspection.

This paper argues that automation should be elevated to a system primitive. We present \textbf{\name}, a new \emph{AgentOS} for Windows that addresses these limitations by embedding automation as a deeply integrated OS abstraction—exposing GUI controls, application APIs, and task orchestration as programmable, inspectable, and composable system services.

\section{System Design of \name}
\begin{figure}[t]
    \centering
    \includegraphics[width=\columnwidth]{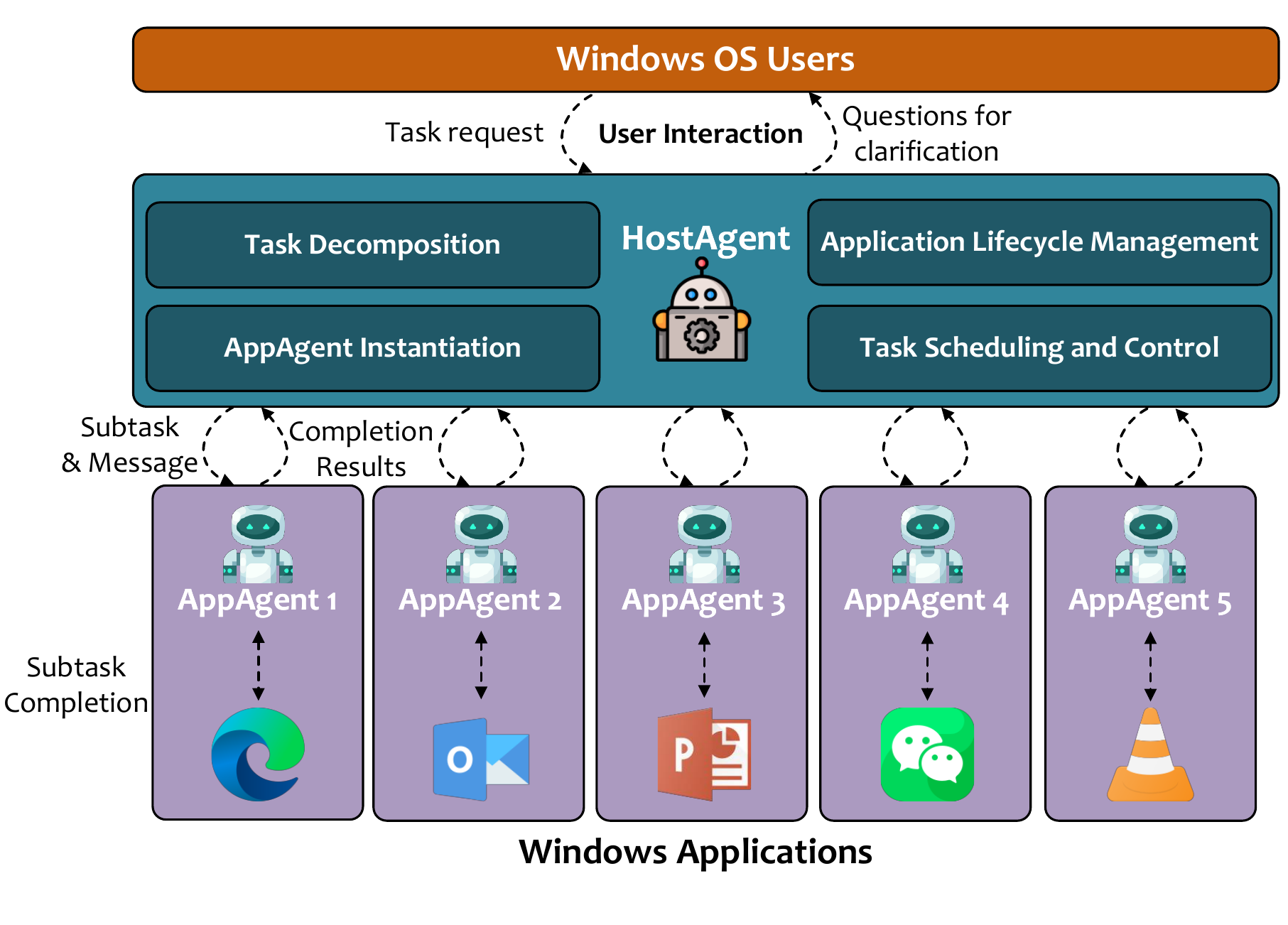}
    \vspace{-3.em}
    \caption{An overview of the architecture of \name.}
    \label{fig:framework}
    \vspace{-1em}
\end{figure}
Motivated by the challenges highlighted in Section~\ref{sec:background}, \name is designed to seamlessly interpret natural language user requests and reliably automate tasks across a wide range of Windows applications. This section provides an architectural overview of \name (Section~\ref{sec:system:nutshell}) and explains how each component is deeply integrated with the underlying OS to overcomes the pitfalls of current CUAs, ultimately enabling a practical, robust AgentOS for desktop automation.

\subsection{\name as a System Substrate for Automation}
\label{sec:system:nutshell}

Figure~\ref{fig:framework} presents the high-level architecture of \name, which provides a structured runtime environment for task-oriented automation on Windows desktops. Deployed as a local daemon, \name enables users to issue natural language requests that are translated into coordinated workflows spanning multiple GUI applications. The system provides core abstractions for orchestration, introspection, control execution, and agent collaboration—exposing these as system-level services analogous to those in traditional OSes.

At the heart of \name is a central control plane, the \hosta, responsible for parsing user intent, managing system state, and dispatching subtasks to a collection of specialized runtime modules called \appas. Each \appa is dedicated to a particular application (\eg Excel, Outlook, File Explorer) and encapsulates all logic needed to observe and control that application, including API bindings, UI detectors, and knowledge bases. These modules act as isolated execution contexts with application-specific semantics.

Upon receiving a user request, \hosta decomposes it into a series of subtasks, each mapped to the application best suited to fulfill it. If the corresponding application is not already running, \hosta launches it using native Windows APIs and instantiates the corresponding \appa. Execution proceeds through a structured loop: each \appa continuously observes the application state (via accessibility APIs and vision-based detectors), reasons about the next operation using a ReAct-style planning cycle~\cite{yao2023react}, and invokes the appropriate action—either a GUI event or a native API call. This loop continues until the subtask terminates, either successfully or due to an unrecoverable error.

\name implements shared memory and control flow via a global blackboard interface, allowing \hosta and \appas to exchange intermediate results, dependency states, and execution metadata. This architecture supports complex workflows across application boundaries—for instance, extracting data from a spreadsheet and using it to populate fields in a web form—without requiring hand-crafted scripts or coordination logic. Crucially, all interactions occur within a virtualized, PiP-based desktop environment, ensuring process-level isolation and safe multi-application concurrency.

\paragraph{Design Rationale: Centralized Multiagent Runtime.}

\name adopts a centralized multiagent~\cite{zhang2024ufo, qiao2023taskweaver, han2024llm, wang2024survey} runtime to support both reliability and extensibility. The centralized \hosta acts as a control plane, simplifying task-level orchestration, error handling, and lifecycle management. Meanwhile, each \appa is architected as a loosely coupled executor that encapsulates deep, application-specific functionality. 

Modularity at the \appa level allows developers and third-party contributors to incrementally expand \name's capabilities by authoring new application interfaces and API bindings. These agents are discoverable, self-contained, and dynamically instantiated by the runtime as needed. From a security and evolvability perspective, this separation of concerns ensures that application logic can evolve independently of the core task orchestration engine.

Together, the \hosta–\appa model allows \name to function as a scalable, pluggable runtime substrate for GUI automation—abstracting away the complexity of heterogeneous interfaces and providing a unified system interface to structured application behavior.

\subsection{\hosta: System-Level Orchestration and Execution Control}
\label{sec:hosta}

\begin{figure}[t]
    \centering
    \includegraphics[width=\columnwidth]{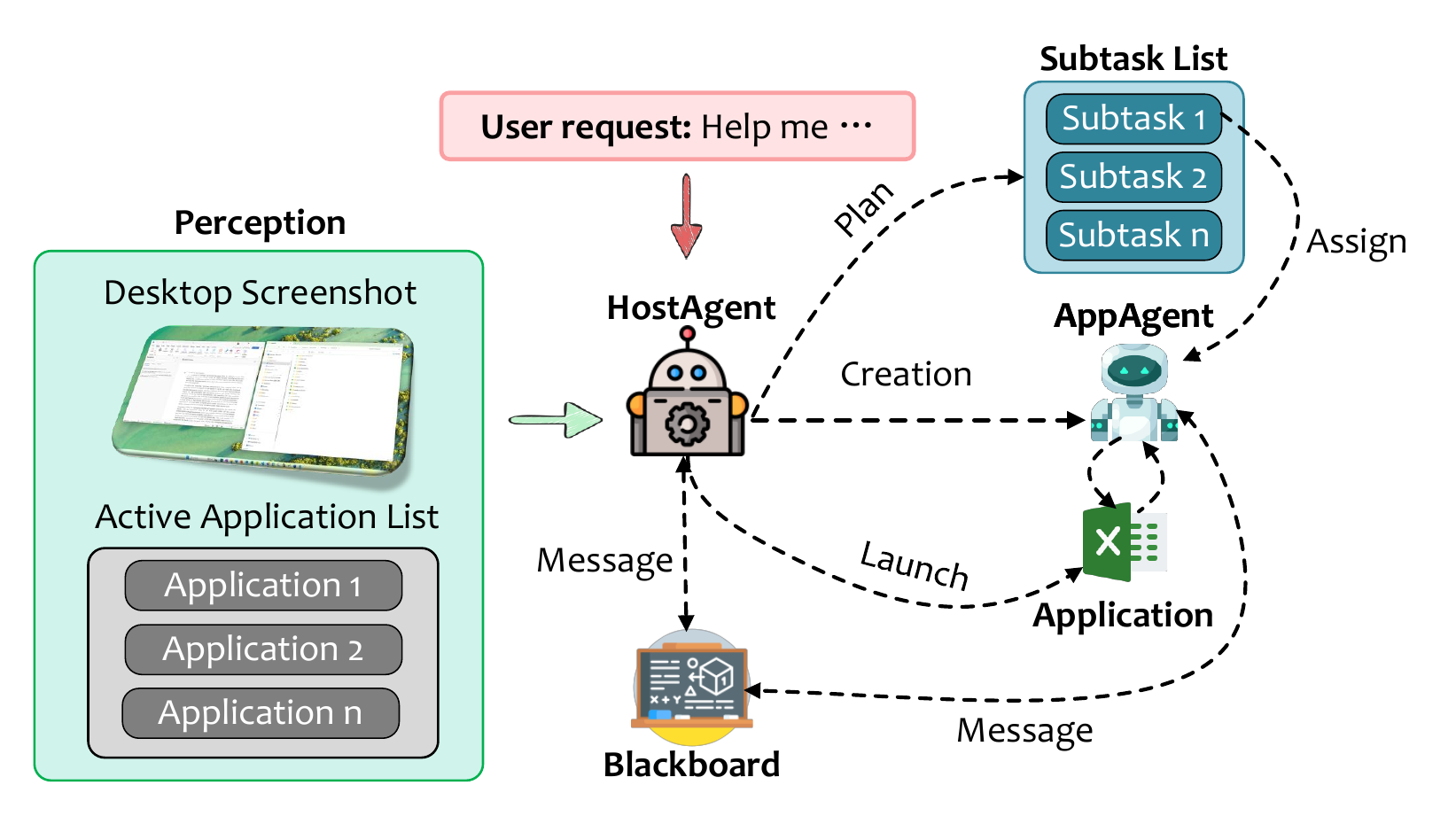}
    \vspace{-2.em}
    \caption{High-level architecture of \hosta as a control-plane orchestrator.}
    \label{fig:hostagent}
    \vspace{-1em}
\end{figure}

The \hosta serves as the centralized control plane of \name. It is responsible for interpreting user-specified goals, decomposing them into structured subtasks, instantiating and dispatching \appa modules, and coordinating their progress across the system. \hosta provides system-level services for introspection, planning, application lifecycle management, and multi-agent synchronization.

Figure~\ref{fig:hostagent} outlines the architecture of \hosta. Operating atop the native Windows substrate, \hosta monitors active applications, issues shell commands to spawn new processes as needed, and manages the creation and teardown of application-specific \appa instances. All coordination occurs through a persistent state machine, which governs the transitions across execution phases.

\paragraph{Responsibilities and Interfaces.}
\hosta exposes the following system services:

\begin{itemize}
    \item \textbf{Task Decomposition.} Given a user's natural language input, \hosta identifies the underlying task goal and decomposes it into a dependency-ordered subtask graph.
    
    \item \textbf{Application Lifecycle Management.} For each subtask, \hosta inspects system process metadata (via UIA APIs) to determine whether the target application is running. If not, it launches the program and registers it with the runtime.

    \item \textbf{\appa Instantiation.} \hosta spawns the corresponding \appa for each active application, providing it with task context, memory references, and relevant toolchains (\eg APIs, documentation).

    \item \textbf{Task Scheduling and Control.} The global execution plan is serialized into a finite state machine (FSM), allowing \hosta to enforce execution order, detect failures, and resolve dependencies across agents.

    \item \textbf{Shared State Communication.} \hosta reads from and writes to a global blackboard, enabling inter-agent communication and system-level observability for debugging and replay.
\end{itemize}

\paragraph{System Perception and Introspection.}
To perform its control functions, \hosta fuses two layers of system introspection:
\begin{enumerate}
    \item \textbf{Visual Layer.} Captures pixel-level screenshots of the desktop workspace, enabling coarse-grained layout understanding.
    \item \textbf{Semantic Layer.} Queries Windows UIA APIs to extract structural metadata about applications, windows, and control hierarchies.
\end{enumerate}
This dual perception enables \hosta to resolve ambiguities, detect runtime inconsistencies, and guide agents with context-aware decisions.

\paragraph{Structured Output Interface.}
\hosta produces structured outputs to drive downstream execution:
\begin{itemize}
    \item \textit{Subtask Plan:} A high-level execution plan detailing decomposed subtasks.
    \item \textit{Shell Command:} A sequence of shell-level invocations for managing application lifecycles.
    \item \textit{Assigned Application:} The process name and index of the application selected for instantiating the \appa, which will be used to execute the next subtask.
    \item \textit{Agent Messages:} Context-specific instructions passed to \appa instances for localized execution.
    \item \textit{User Prompts:} Interactive clarification requests in cases of ambiguity or failure.
    \item \textit{\hosta State:} The current state within the \hosta's internal FSM.
\end{itemize}

\paragraph{Execution via Finite-State Controller.}
\begin{figure}[t]
    \centering
    \includegraphics[width=\columnwidth]{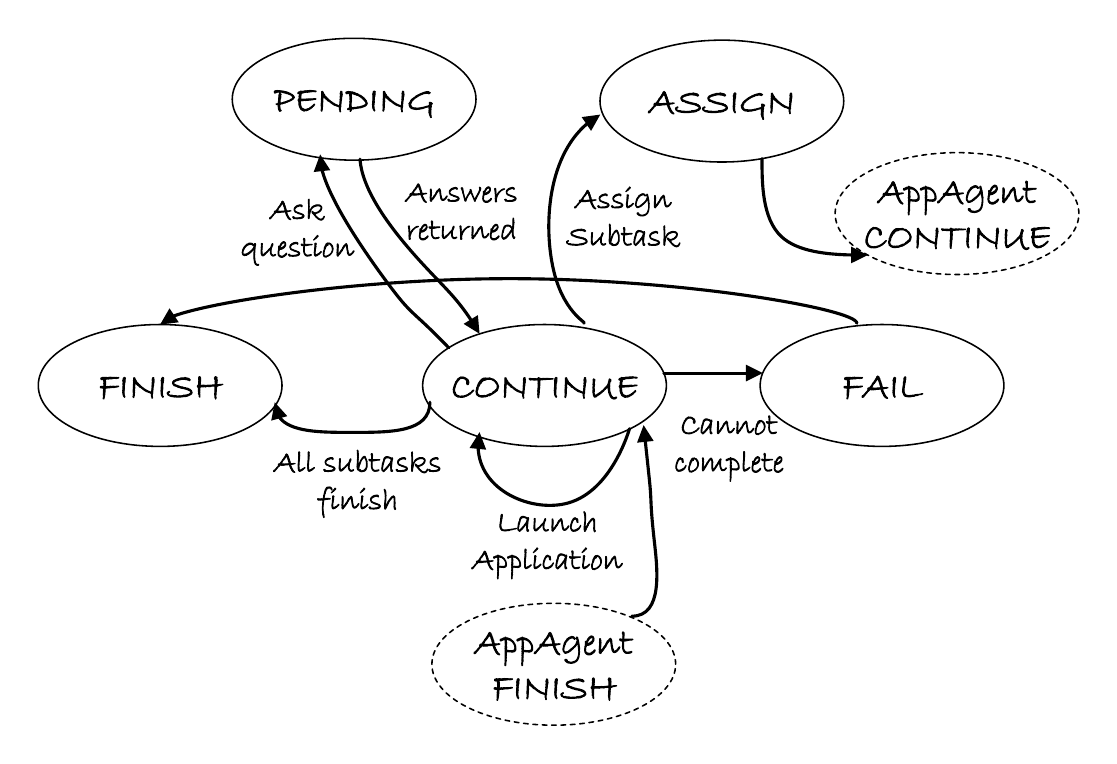}
    \vspace{-2.5em}
    \caption{Control-state transitions managed by \hosta.}
    \label{fig:host_state}
    \vspace{-1em}
\end{figure}

The core logic of \hosta is modeled as a finite-state controller (Figure~\ref{fig:host_state}), with the following states:
\begin{itemize}
    \item \texttt{CONTINUE:} Main execution loop; evaluates which subtasks are ready to launch or resume.
    \item \texttt{ASSIGN:} Selects an available application process and spawns the corresponding \appa agent.
    \item \texttt{PENDING:} Waits for user input to resolve ambiguity or gather additional task parameters.
    \item \texttt{FINISH:} All subtasks complete; cleans up agent instances and finalizes session state.
    \item \texttt{FAIL:} Enters recovery or abort mode upon irrecoverable failure.
\end{itemize}
This explicit FSM structure enables \hosta to robustly orchestrate dynamic workflows while maintaining high-level guarantees over task completion and fault isolation.

\paragraph{Memory and State Management.}
\hosta maintains two classes of persistent state:
\begin{itemize}
    \item \textbf{Private State:} Tracks user intent, plan progress, and the control flow of the current session.
    \item \textbf{Shared Blackboard:} A concurrent, append-only memory space that facilitates transparent agent communication by recording key observations, intermediate results, and execution metadata accessible to all \appa instances.
\end{itemize}
This separation ensures that local context remains encapsulated, while global coordination is visible and consistent across the system.This separation ensures that each agent retains clean, scoped state while benefiting from a globally consistent view for collaborative task execution.

Overall, \hosta abstracts away the complexity of managing concurrent, stateful, cross-application workflows in desktop environments \cite{zhang2024survey}. Its control-plane role enables modular execution, coordinated progress, and robust task lifecycle management—all critical features in scaling desktop automation to real-world deployments.

\subsection{\appa: Application-Specialized Execution Runtime}
\label{sec:appa}

\begin{figure*}[t]
    \centering
    \includegraphics[width=0.8\textwidth]{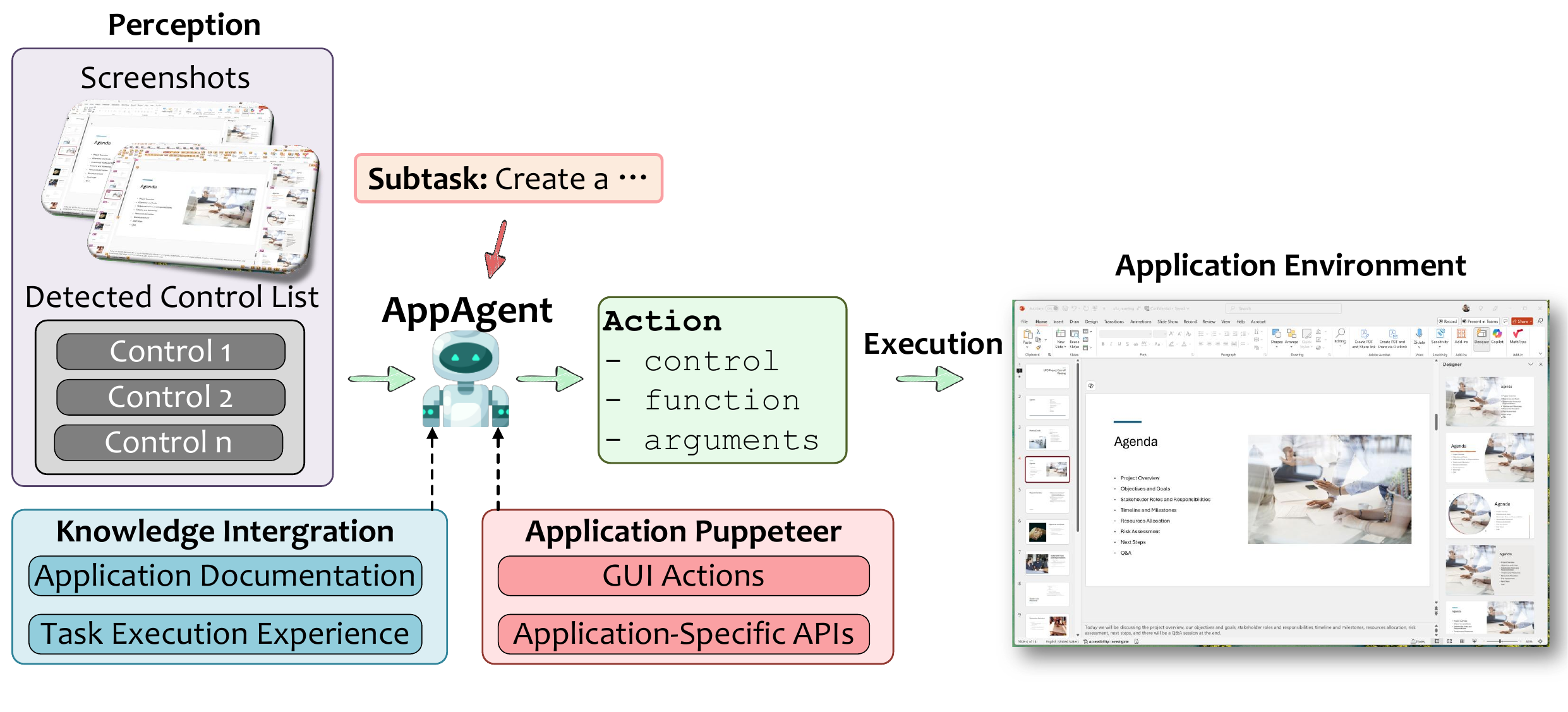}
    \vspace{-1.5em}
    \caption{Architecture of an \appa, the per-application execution runtime in \name.}
    \label{fig:appagent}
    \vspace{-1em}
\end{figure*}

The \appa is the core execution runtime in \name, responsible for carrying out individual subtasks within a specific Windows application. Each \appa functions as an isolated, application-specialized worker process launched and orchestrated by the central \hosta (Section~\ref{sec:hosta}). Unlike monolithic CUAs that treat all GUI contexts uniformly, each \appa is tailored to a single application and operates with deep knowledge of its API surface, control semantics, and domain logic.

Figure~\ref{fig:appagent} outlines the architecture of an \appa. Upon receiving a subtask and execution context from the \hosta, the \appa initializes a ReAct-style control loop~\cite{yao2023react}, where it iteratively senses the current application state, reasons about the next step, and executes either a GUI or API-based action. This hybrid execution layer—implemented via a \texttt{Puppeteer} interface—enables reliable control over dynamic and complex UIs by favoring structured APIs whenever available, while retaining fallback to GUI-based interaction when necessary.

\paragraph{Perception Layer.}
Each \appa fuses multiple streams of perception:
\begin{itemize}
    \item \textbf{Visual Input:} Captures GUI screenshots for layout understanding and control grounding.
    \item \textbf{Semantic Metadata:} Extracted from Windows UIA APIs, including control type, label, hierarchy, and enabled state.
    \item \textbf{Symbolic Annotation:} Uses Set-of-Mark (SoM) techniques~\cite{yang2023set} to annotate the control on screenshots.
\end{itemize}
These fused signals are converted into a structured observation object containing both the GUI screenshot and the set of candidate control elements. This multi-modal representation enables a more comprehensive understanding of the application state, going well beyond raw visual input alone.

\paragraph{Structured Output.}
Based on this state, the \appa produces a structured output:
\begin{itemize}
    \item Target control (if applicable)
    \item Action type (\eg click, type, call API)
    \item Arguments or payload
    \item Reasoning trace and planning with Chain-of-Thought (CoT)~\cite{wei2022chain, ding2024everything}
    \item Current state in local FSM
\end{itemize}
This design decouples perception from actuation, enabling deterministic replay, offline debugging, and fine-grained observability.

\paragraph{Execution via Finite-State Controller.}
\begin{figure}[t]
    \centering
    \includegraphics[width=\columnwidth]{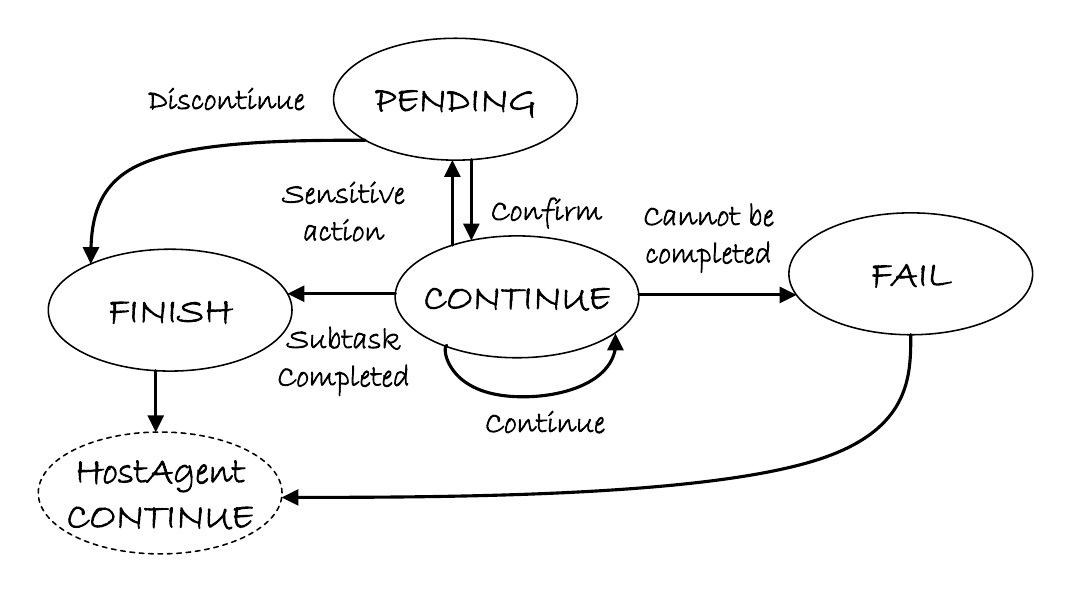}
    \vspace{-2.5em}
    \caption{Control-state transitions for an \appa runtime.}
    \label{fig:app_state}
    \vspace{-1em}
\end{figure}

Each \appa maintains a local finite-state machine (Figure~\ref{fig:app_state}) that governs its behavior within the assigned application context:
\begin{itemize}
    \item \texttt{CONTINUE:} Default state for action planning and execution.
    \item \texttt{PENDING:} Invoked for safety-critical actions (\eg destructive operations); requires user confirmation.
    \item \texttt{FINISH:} Task completed; execution ends.
    \item \texttt{FAIL:} Irrecoverable failure detected (\eg application crash, permission error).
\end{itemize}

This bounded execution model isolates failures to the current task and enables safe preemption, retry, or delegation. The FSM also supports interruptible workflows, which can resume from intermediate checkpoints.

\begin{figure*}[t]
    \centering
    \includegraphics[width=\textwidth]{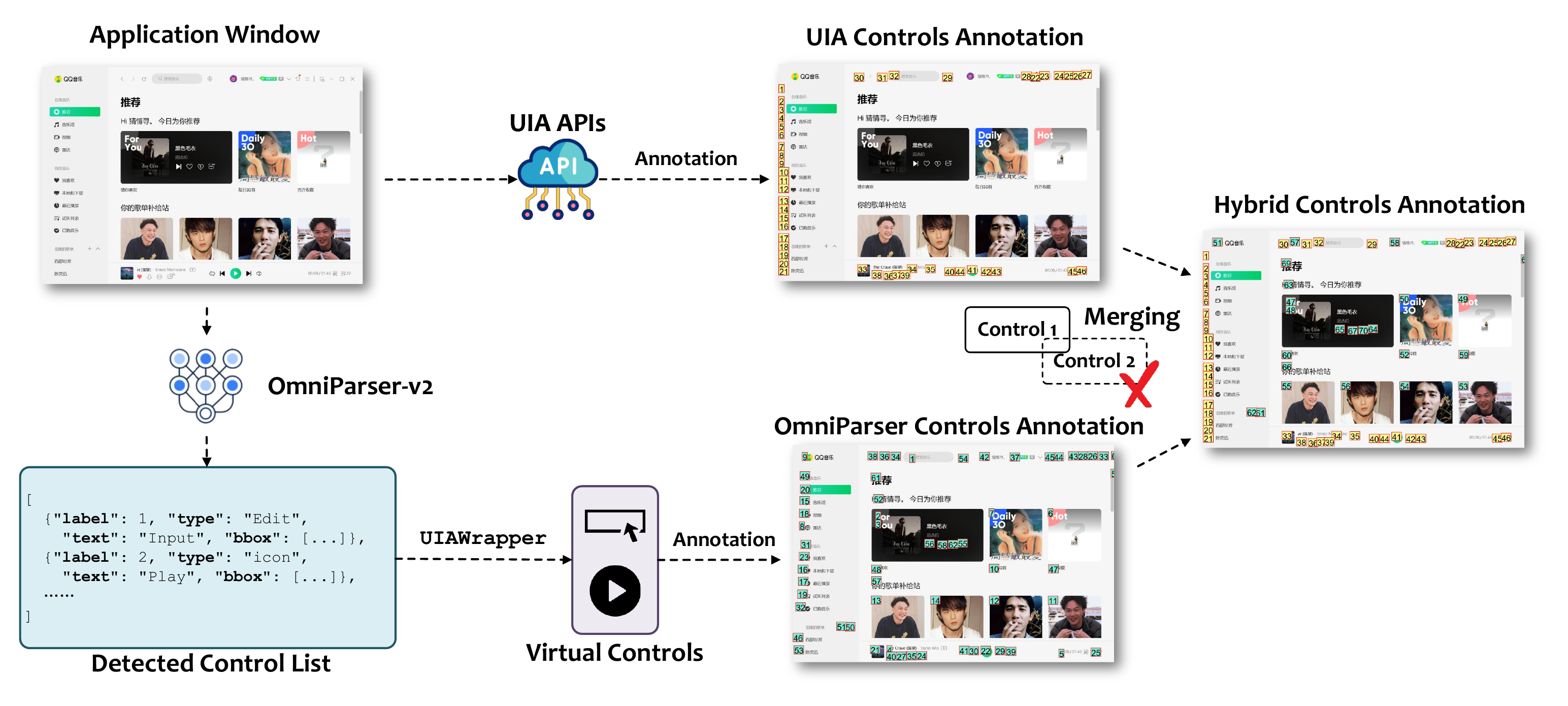}
    \vspace{-2.5em}
    \caption{The hybrid control detection approach employed in \name.}
    \label{fig:controls}
\end{figure*}

\paragraph{Memory and State Coordination.}
To enable stateful execution and maintain contextual awareness, each \appa maintains:
\begin{itemize}
    \item \textbf{Private State:} A local log of all executed actions, control decisions, and CoT traces.
    \item \textbf{Shared State:} Updates to the system-wide blackboard, including intermediate outputs, encountered errors, and application-level insights.
\end{itemize}

This dual-memory design enables \appas to act autonomously on behalf of \hosta while remaining synchronized with the broader system. It also supports composability: one \appa's output may become another's input in downstream subtasks.

\paragraph{Application-Aware SDK and Extensibility.}
To support rapid onboarding of new applications, \name exposes an SDK that encapsulates the development and maintenance of \appas. Developers can register application-specific APIs via a declarative interface, including function metadata, argument schemas, and prompt bindings. Domain-specific help documents and patch notes can be ingested into a searchable knowledge base that agents query at runtime.

This modular abstraction allows third-party vendors or power users to extend \name's capabilities without retraining any models. New functionality can be integrated by updating the application's \appa module, isolating changes from the rest of the system and minimizing regression risk.

\paragraph{Summary.}
As a per-application execution runtime, each \appa provides modular, domain-aware control that surpasses generic GUI agents in both efficiency and robustness. Its hybrid perception-action loop, plugin-based extensibility, and local fault containment enable \name to scale to large application ecosystems with minimal system-wide disruption.

\subsection{Hybrid Control Detection}
\label{sec:hybrid-control}
Reliable perception of GUI elements is fundamental to enabling \appas to interact with application interfaces in a deterministic and safe manner. However, real-world GUI environments exhibit substantial heterogeneity: some applications expose well-structured accessibility data via Windows UI Automation (UIA) APIs, while others—especially legacy or custom applications—render critical controls using non-standard toolkits that bypass UIA entirely.

To address this disparity, \name introduces a \emph{hybrid control detection} subsystem that fuses UIA-based metadata with vision-based grounding~\cite{lu2024omniparser, cheng2024seeclick, gou2024navigating} to construct a unified and comprehensive control graph for each application window (Figure~\ref{fig:controls}). This design ensures both coverage and reliability, forming a resilient perceptual foundation for downstream action planning and execution.

\paragraph{UIA-Layer Detection.}
When available, UIA offers a semantically rich and high-precision interface for enumerating on-screen controls. The detection pipeline first queries the accessibility tree to extract controls satisfying a set of runtime predicates (\eg \texttt{is\_visible()}, \texttt{is\_enabled()}). These controls are annotated with their attributes (type, label, bounding box) and assigned stable identifiers, forming the initial control graph.

\paragraph{Vision-Layer Augmentation.}
To augment the perception pipeline for UI-invisible or custom-rendered controls, we integrate OmniParser-v2~\cite{lu2024omniparser}, a vision-based grounding model designed for fast and accurate GUI parsing. OmniParser-v2 combines a lightweight YOLO-v8~\cite{reis2023real} detector with a fine-tuned Florence-2 (0.23B)~\cite{xiao2024florence} encoder to process raw application screenshots and identify additional interactive elements. Each detection includes the control type, confidence score, and spatial bounding box.

\paragraph{Fusion and Deduplication.}
We unify these two streams by performing deduplication based on bounding-box overlap. Visual detections with Intersection-over-Union (IoU) greater than 10\% against any UIA-derived control are discarded. Remaining visual-only detections are converted into pseudo-UIA objects using a lightweight \texttt{UIAWrapper} abstraction, allowing them to seamlessly integrate into the rest of the \appa pipeline. This fused control set is passed downstream to the SoM-based annotation module~\cite{yang2023set}. Figure~\ref{fig:controls} shows a typical scenario involving a hybrid-rendered GUI. Yellow bounding boxes denote standard UIA-detected elements, while blue bounding boxes represent visual-only detections. Both are integrated into a single actionable control graph consumed by the \appa.

\subsection{Unified GUI--API Action Orchestrator}
\label{sec:api}

\begin{figure}[t]
    \centering
    \includegraphics[width=\columnwidth]{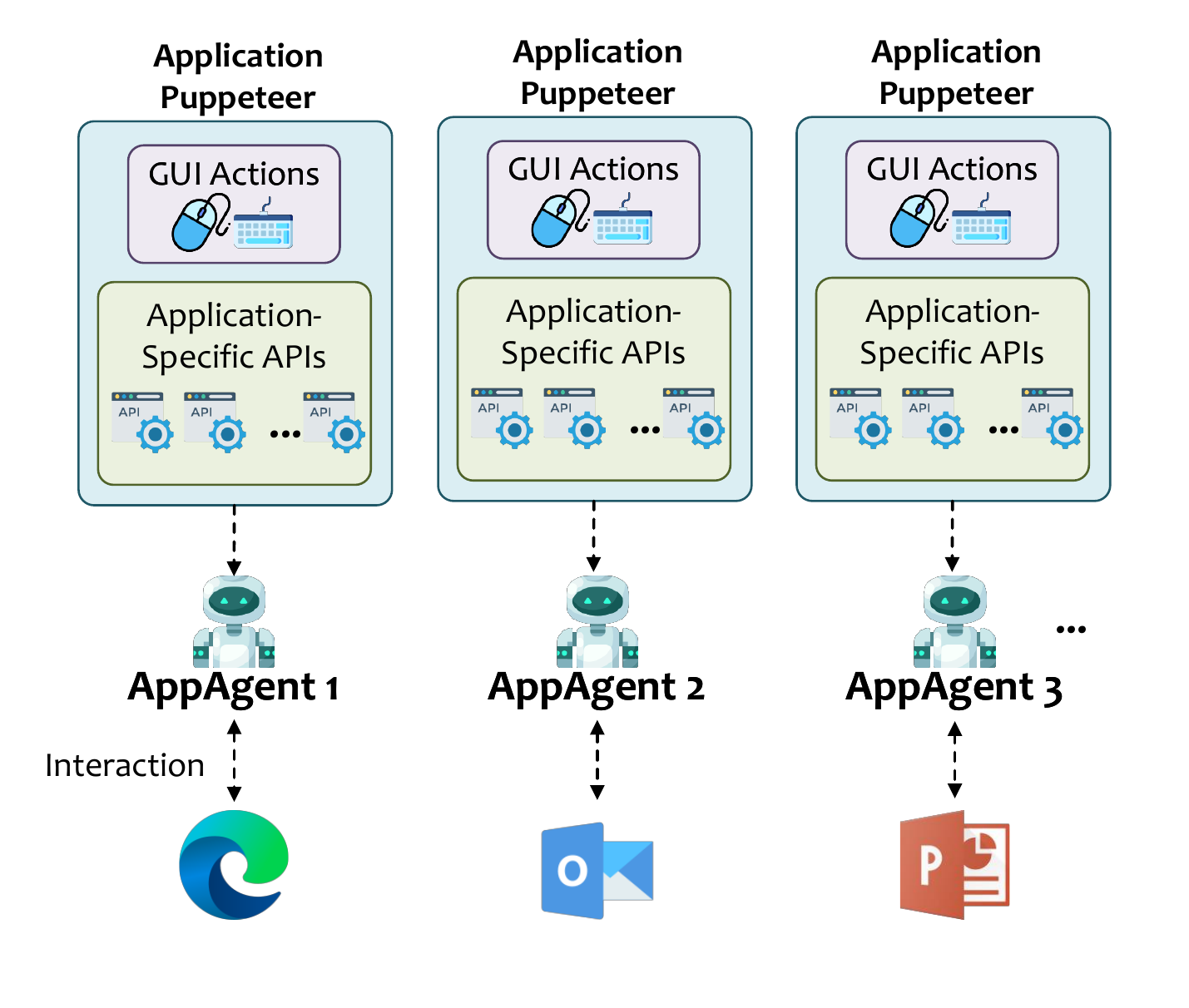}
    \vspace{-3.5em}
    \caption{\texttt{Puppeteer} serves as a unified execution engine that harmonizes GUI actions and native API calls.}
    \label{fig:puppeteer}
    \vspace{-1em}
\end{figure}

\appas can interact with application environments that expose two distinct classes of interfaces: GUI frontends, which are universally observable but often brittle; and native APIs, which are high-fidelity but require explicit integration \cite{zhang2025apiagentsvsgui}. To unify these heterogeneous execution backends under a single runtime abstraction, \name introduces the \texttt{Puppeteer}, a modular execution orchestrator that dynamically selects between GUI-level automation and application-specific APIs for each action step (Figure~\ref{fig:puppeteer}). This design significantly improves task robustness, latency, and maintainability. Tasks that would otherwise require long GUI interaction sequences (\eg iteratively selecting and formatting cells in Excel) can often be collapsed into a single API call, reducing both execution time and the surface area for failure~\cite{song2024beyond, zhang2025apiagentsvsgui}.

\texttt{Puppeteer} supports a lightweight API registration mechanism that enables developers to expose high-level operations in target applications. APIs are registered using a simple Python decorator interface, as shown in Figure~\ref{fig:api}. Each function is wrapped with metadata—name, argument schema, and application binding—and automatically incorporated into the \appa's runtime action space.

\begin{figure}[t]
\centering
\begin{minipage}{\columnwidth}
\begin{lstlisting}[style=pythonstyle, label={lst:greet}]
@ExcelWinCOMReceiver.register
class SelectTableRangeCommand:
    def execute(self) -> Dict[str, str]:
        ...
        return {"results":... "error":...}
    @classmethod
    def name(cls) -> str:
        return "select_table_range"
\end{lstlisting}
\end{minipage}
\caption{Example API registration for Excel.}
\label{fig:api}
\end{figure}

At runtime, \name prompts \appa to employs a decision-making policy to choose the most appropriate execution path for each operation. If a semantically equivalent API is available, \texttt{Puppeteer} prefers it over GUI automation for reliability and atomicity.If an API fails or is unavailable (\eg missing bindings or runtime permission errors), the system gracefully falls back to GUI-based control via simulated clicks or keystrokes. This runtime flexibility allows \appa to preserve robustness across heterogeneous environments without sacrificing generality.

\texttt{Puppeteer} transforms action execution in \name from a monolithic GUI-centric model into a flexible, OS-integrated control layer that mixes perceptual agility with semantic precision. This hybrid execution model not only improves system performance and stability but also lays the groundwork for sustainable integration of application-specific capabilities in future desktop agents.

\subsection{Continuous Knowledge Integration Substrate}
\label{sec:knowledge}

\begin{figure}[t]
    \centering
    \includegraphics[width=\columnwidth]{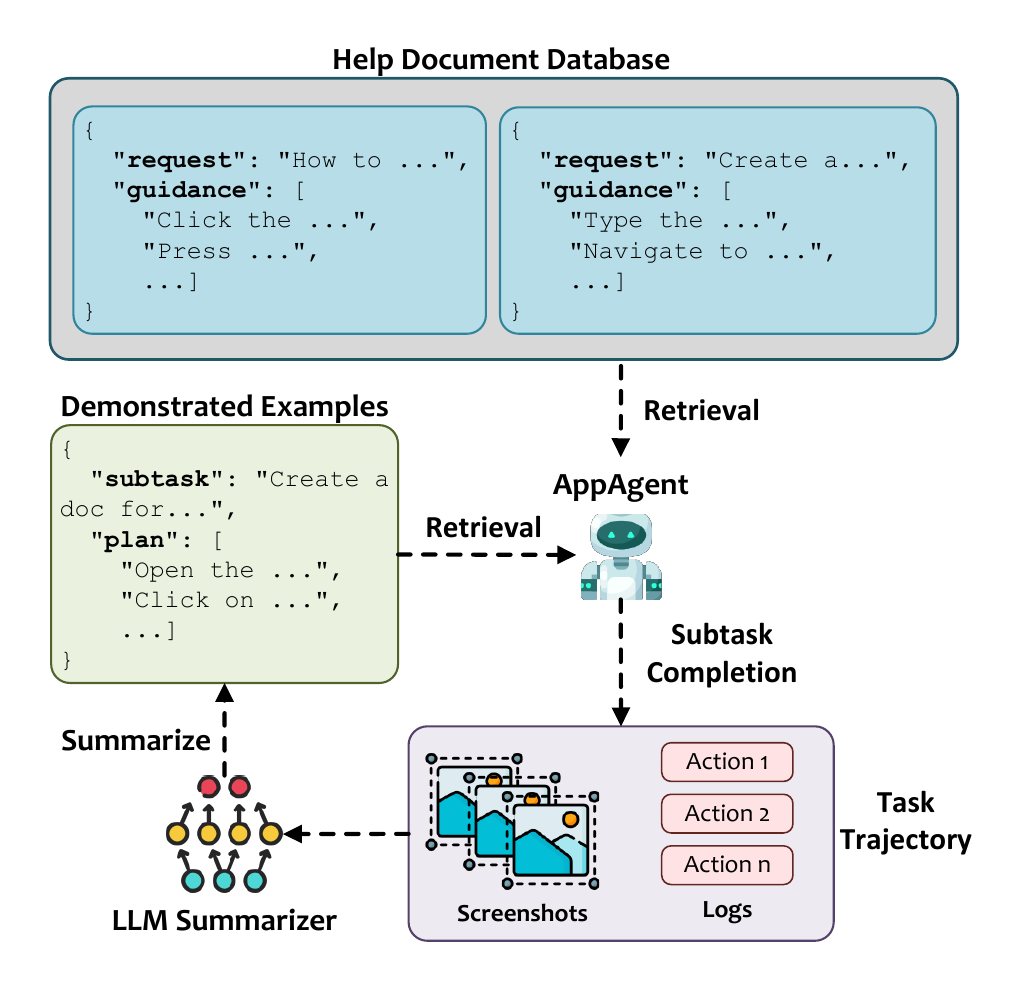}
    \vspace{-3em}
    \caption{Overview of the knowledge substrate in \name, combining static documentation with dynamic execution history.}
    \label{fig:knowledge}
    \vspace{-1em}
\end{figure}

Unlike traditional CUAs, which rely heavily on static training corpora, \name introduces a persistent and extensible \emph{knowledge substrate} that supports runtime augmentation of application-specific understanding. As illustrated in Figure~\ref{fig:knowledge}, this substrate enables each \appa to retrieve, interpret, and apply external documentation and prior execution traces without requiring retraining of the underlying models. This hybrid memory design functions analogously to an OS-level metadata manager, abstracting over two key knowledge flows: static references (\eg user manuals) and dynamic experience (\eg execution logs).

\paragraph{Bootstrapping from Documentation.}
Most real-world desktop applications expose substantial task-level documentation via user guides, help menus, or online tutorials. \name capitalizes on this resource by offering a one-click interface to parse and ingest such documents into an application-specific vector store. Documents are structured as \texttt{json} records with a natural language description in the \texttt{request} field and detailed execution guidance in the \texttt{guidance} field.

At runtime, when an \appa receives a subtask, it queries this indexed store to retrieve relevant guidance, which is then injected into the agent's prompt. This mechanism effectively mitigates cold-start issues—especially when handling novel applications or infrequent operations—by enriching the agent's reasoning context with domain-grounded procedural knowledge.

\paragraph{Reinforcing from Experience.}
Beyond static knowledge, \name continuously learns from its own execution history. Each automation run produces structured logs—including natural language task descriptions, executed action sequences, application screenshots, and final outcomes. Periodically, these logs are mined offline by a summarization module that distills successful trajectories into reusable \texttt{Example} records.

Each record contains a task signature and an associated step-by-step plan, stored in an application-specific example database. When a similar task is encountered in the future, \appa uses In-Context Learning (ICL)~\cite{dong2024survey, min2022rethinking, zhang2024allhands, luocontext} to retrieve relevant demonstrations and improve execution fidelity. This dynamic reinforcement pipeline transforms the system into a long-lived agent that improves with use, without introducing the brittleness or operational cost of fine-tuning~\cite{wang2025benchmark}.

\paragraph{Runtime RAG Integration.}
At the system level, the knowledge substrate acts as a Retrieval-Augmented Generation (RAG) layer~\cite{lewis2020retrieval, liu2024large, gao2023retrieval, jiang2024xpert} that bridges the gap between pre-trained language models and application-specific requirements. Because both help documents and examples are indexed with semantic embeddings, the retrieval pipeline is fast, interpretable, and cache-friendly. Additionally, versioned indexing ensures that knowledge artifacts can evolve alongside software updates, preventing model obsolescence and supporting robust execution across long deployment cycles.

By integrating static and experiential knowledge into a unified RAG pipeline, \name transforms CUAs from brittle, training-time constructs into dynamic, evolving agents. This substrate plays a foundational role in enabling sustainable automation across complex, heterogeneous application ecosystems.

\subsection{Speculative Multi-Action Execution\label{sec:speculative}}
\label{sec:multiaction}

\begin{algorithm}[t]
  \caption{Speculative Multi‑Action Execution in \name}
  \label{alg:speculative-exec}
  \begin{algorithmic}[1]
    \Require Initial UI context $C_0$, batch size $k$
    \Ensure  List \texttt{Executed} of actions completed so far
    %
    \Statex\textbf{Stage 1: Batch Prediction}
    \State $A \gets \textsc{LLM\_Predict}(C_0, k)$
           \Comment{$A=\bigl[(\mathit{ctrl}_i,\mathit{op}_i\bigr]_{i=1}^k$}
    \State \texttt{Executed} $\gets [\ ]$;
           \quad $C \gets C_0$
    %
    \Statex\textbf{Stage 2 \& 3: Sequential Validate‑Execute Loop}
    \For{$i \gets 1$ \textbf{to} $k$}
        \State $(ctrl, op, \_) \gets A[i]$
        \Statex\quad\textit{// Validate in the \emph{current} context}
        \If{\textbf{not} \textsc{UIA\_IsEnabled}$(ctrl,C)\ \lor$
             \textbf{not} \textsc{UIA\_IsVisible}$(ctrl,C)$}
            \State \textbf{break} \Comment{validation failed $\rightarrow$ early stop}
        \EndIf
        \Statex\quad\textit{//Execute and refresh context}
        \State \textsc{Execute}$(ctrl,op)$
        \State \texttt{append} $(ctrl,op)$ to \texttt{Executed}
        \State $C \gets \textsc{UIA\_GetContext}()$  \Comment{UI changed}
    \EndFor
    %
    \If{$| \texttt{Executed} | < k$}
        \State \textsc{ReportPartial}(\texttt{Executed})
        \State \textsc{Replan}$(C)$
    \EndIf
  \end{algorithmic}
\end{algorithm}

\begin{figure*}[t]
    \centering
    \includegraphics[width=\textwidth]{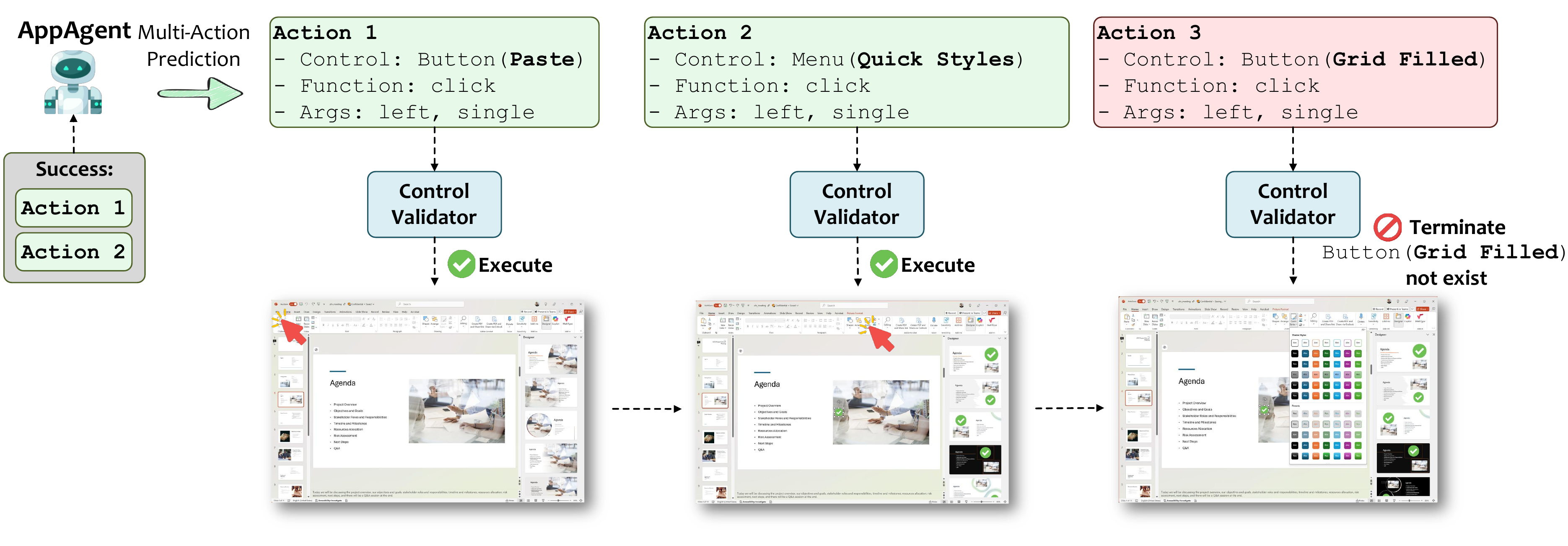}
    \vspace{-2.5em}
    \caption{Speculative multi-action execution in \name: batched inference with online validation.}
    \label{fig:multiaction}
\end{figure*}

Conventional CUAs suffer from a fundamental execution bottleneck: each automation step is executed in isolation, requiring a full LLM inference for every single GUI action. This \emph{step-wise inference loop} introduces excessive latency, inflates system resource usage, and increases cumulative error rates—especially when interacting with complex or multi-phase workflows~\cite{zhang2025apiagentsvsgui}. The root cause is the dynamic and uncertain nature of GUI environments, where any single action may alter the interface and invalidate future plans.

To overcome these limitations, \name introduces a system-level optimization called \emph{speculative multi-action execution}, inspired by classical ideas from speculative execution in processor design and instruction pipelining. Rather than issuing one action per LLM call, \name speculatively generates a \emph{batch} of likely next steps using a single inference pass and validates their applicability at runtime through tight OS integration. We present an algorithm in~\ref{alg:speculative-exec}.

The speculative executor operates in three stages:
\begin{enumerate}
    \item \textbf{Action Prediction:} The \appa issues a single LLM query to predict multiple plausible actions under its current context. Each predicted step includes a target control, intended operation, and rationale.
    \item \textbf{Runtime Validation:} For each action, the system consults the Windows UIA API to verify the action's preconditions (\eg \texttt{is\_enabled()}, \texttt{is\_visible()}). This check ensures that each target control is still valid and interactive.
    \item \textbf{Sequential Execution and Early Exit:} Actions are executed in order, halting immediately if any validation fails due to interface change (\eg control no longer exists or is disabled). The executor then reports a partial result set and prompts the agent to replan.
\end{enumerate}
We show an illustrative example of speculative multi-action execution in Figure~\ref{fig:multiaction}. In this case, the \appa initially plans to execute three actions in a single step: clicking \texttt{Paste}, then \texttt{Quick Style}, and finally \texttt{Grid Filled}. However, after the second action, the control validator detects that the control required for the third action (\texttt{Grid Filled}) is no longer present—likely because the GUI layout changed as a result of the previous step. The \texttt{Puppeteer} then terminates execution at that point and returns the partial results. This example highlights how \name safely handles speculative execution by validating each control before acting, ensuring robustness even in the face of dynamic interface changes. 

Overall, this strategy drastically reduces LLM invocation frequency and amortizes the cost of action planning across multiple steps, while preserving the correctness guarantees of per-step validation. Critically, validation is performed by trusted OS-level APIs instead of vision models, ensuring high reliability and eliminating spurious interactions.

\section{Picture-in-Picture Interface}
\label{sec:pip}

\begin{figure*}[t]
    \centering
    \includegraphics[width=\textwidth]{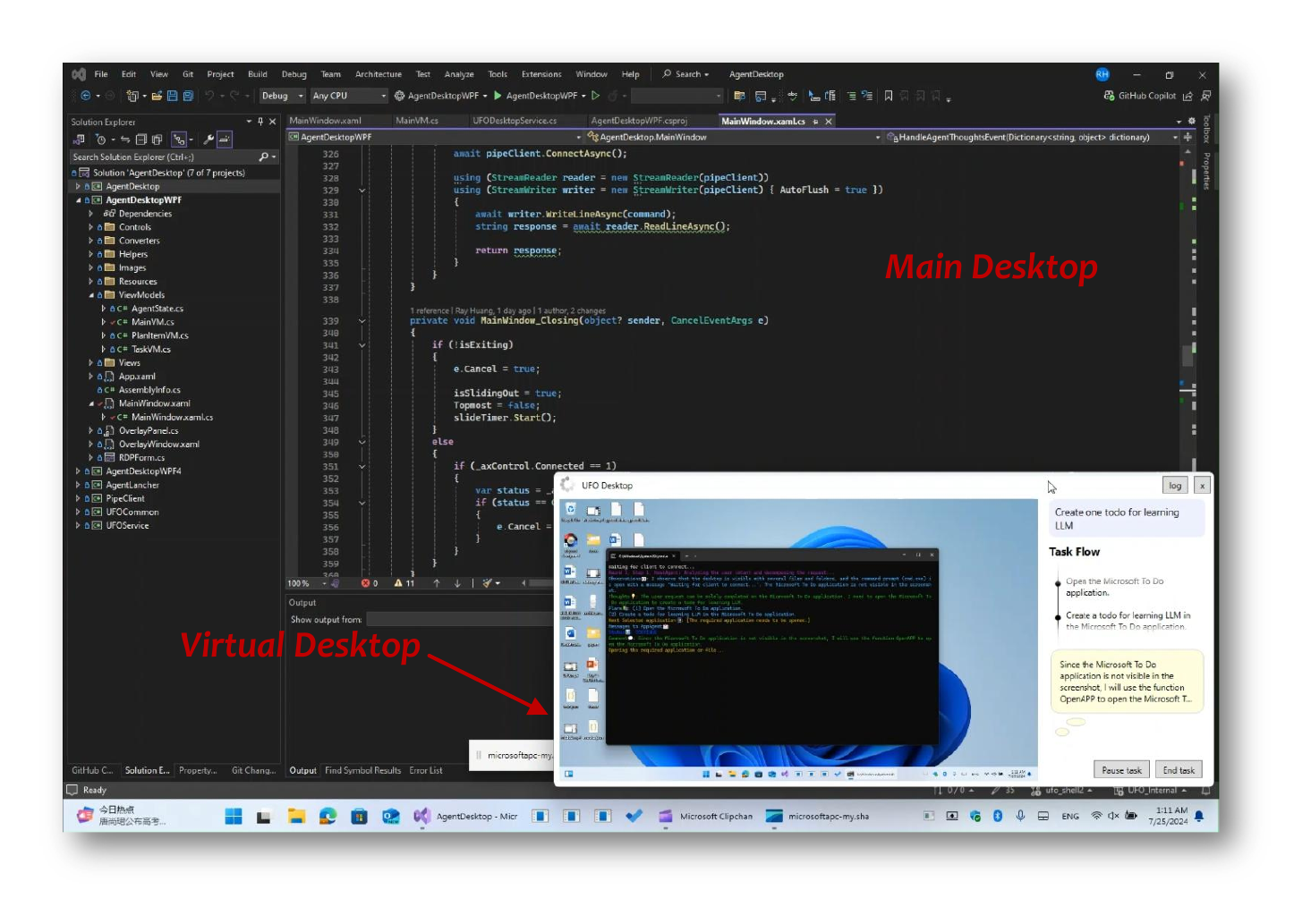}
    \vspace{-4.5em}
    \caption{The Picture-in-Picture interface: a virtual desktop window for non-disruptive automation.}
    \label{fig:pip}
\end{figure*}

A key design objective of \name is to deliver high-throughput automation while preserving the responsiveness and usability of the primary desktop environment. Existing CUAs often monopolize the user's workspace, seizing mouse and keyboard control for extended periods and making the system effectively unusable during task execution. To overcome this, \name introduces a \emph{Picture-in-Picture} (PiP) interface: a lightweight, virtualized desktop window powered by Remote Desktop loopback, enabling fully isolated agent execution in parallel with active user workflows, as illustrated in Figure~\ref{fig:pip}.

\subsection{Virtualized User Environment with Minimal Disruption}

Unlike conventional CUAs that operate in the main desktop session, the PiP interface presents a resizable, movable window containing a fully functional replica of the user's desktop. Internally, this is implemented via Windows' native Remote Desktop Protocol (RDP) loopback \cite{miller2007virtualization}, creating a distinct virtual session hosted on the same machine. Applications launched within the PiP session inherit the user's identity, credentials, settings, and network context, ensuring consistency with foreground operations.

From the user's perspective, the PiP window behaves like a sandboxed workspace: the automation executes in the background, visible but unobtrusive. The user retains full control of the primary desktop and can minimize or reposition the PiP window at will. This enables \name to perform long-running or repetitive workflows (\eg data entry, batch file processing) without blocking user interaction or degrading responsiveness.

\subsection{Robust Input and State Isolation}

To ensure robust separation between agent actions and user activities, \name leverages the RDP subsystem to maintain distinct input queues and device contexts across sessions. Mouse and keyboard events generated within the PiP desktop are fully scoped to that session and cannot interfere with the primary desktop. Similarly, GUI changes and focus transitions are restricted to the virtual environment.

This level of input isolation is critical for preventing accidental interference—either by the user or the agent—and ensures that automation sequences remain stable, even during simultaneous foreground activity. The architecture also supports controlled error recovery: failures or unexpected UI states within the PiP session do not propagate to the primary desktop, preserving the integrity of the user's environment.

\subsection{Secure Cross-Session Coordination}
Although visually and operationally distinct, the PiP session must remain logically connected to the host environment. To enable this, \name establishes a secure inter-process communication (IPC) channel between the PiP agent runtime and a host-side coordinator. We implement this using Windows Named Pipes, authenticated and encrypted using per-session credentials \cite{venkataraman2015evaluation}.

This IPC layer supports two-way messaging:
\begin{itemize}
    \item From the host to the PiP: task assignment, progress polling, cancellation, and user clarifications.
    \item From the PiP to the host: status updates, completion reports, and exception notifications.
\end{itemize}

Users interact with the automation pipeline through a lightweight frontend panel on the host desktop, enabling real-time visibility and partial control without needing to directly access the PiP window. This transparent yet secure communication channel ensures trust and usability, particularly in long-running or partially supervised workflows.

\subsection{System-Level Implications}
The PiP interface represents more than a UX refinement—it is a system-level abstraction that reconciles concurrency, usability, and safety. It decouples automation execution from foreground interactivity, introduces a new isolation primitive for GUI-based agents, and simplifies failure recovery by sandboxing side effects. By exploiting existing RDP capabilities with minimal system overhead, the PiP interface offers a practical and backwards-compatible approach to scalable desktop automation.

\section{Implementation and Specialized Engineering Design}
\label{sec:implementation}

We implement \name as a full-stack desktop automation framework spanning over 30,000 lines of \texttt{Python} and \texttt{C\#} code. Python serves as the core runtime environment for agent orchestration, control logic, and API integration, while \texttt{C\#} supports GUI development, debugging interfaces, and Windows-specific operations such as the Picture-in-Picture desktop. To support retrieval-augmented reasoning, \name leverages Sentence Transformers~\cite{reimers2019sentence} for embedding-based document and experience retrieval.

Beyond its core functionality, \name incorporates multiple specialized engineering components that target critical systems goals: composability, interactivity, debuggability, and scalable deployment. We highlight several key mechanisms below.

\subsection{Multi-Round Task Execution}
\label{sec:session}
\begin{figure}[t]
    \centering
    \includegraphics[width=\columnwidth]{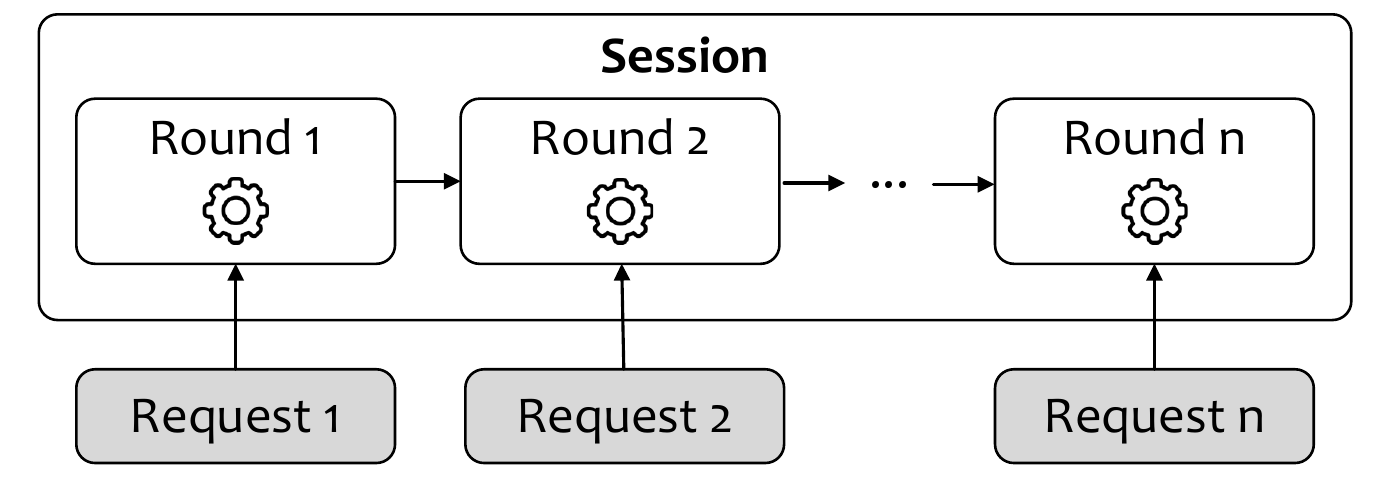}
    \vspace{-2em}
    \caption{The interactive \texttt{Session} model in \name supports multi-round refinement.}
    \label{fig:session}
    \vspace{-1em}
\end{figure}

Unlike stateless one-shot agents, \name adopts a session-based execution model to support iterative, interactive workflows (Figure~\ref{fig:session}). Each \texttt{Session} maintains persistent contextual memory—including intermediate results, task progress, and application state—across multiple \texttt{Rounds} of execution. Users can refine prior instructions, launch follow-up tasks, or intervene when agents encounter ambiguous or unsafe operations.

This multi-round interaction paradigm facilitates progressive convergence on complex tasks while preserving transparency and human oversight. It enables \name to support human-in-the-loop refinement strategies, bridging static LLM workflows with dynamic user guidance.

\subsection{Safeguard Mechanism}
\label{sec:safeguard}

\begin{figure}[t]
    \centering
    \includegraphics[width=\columnwidth]{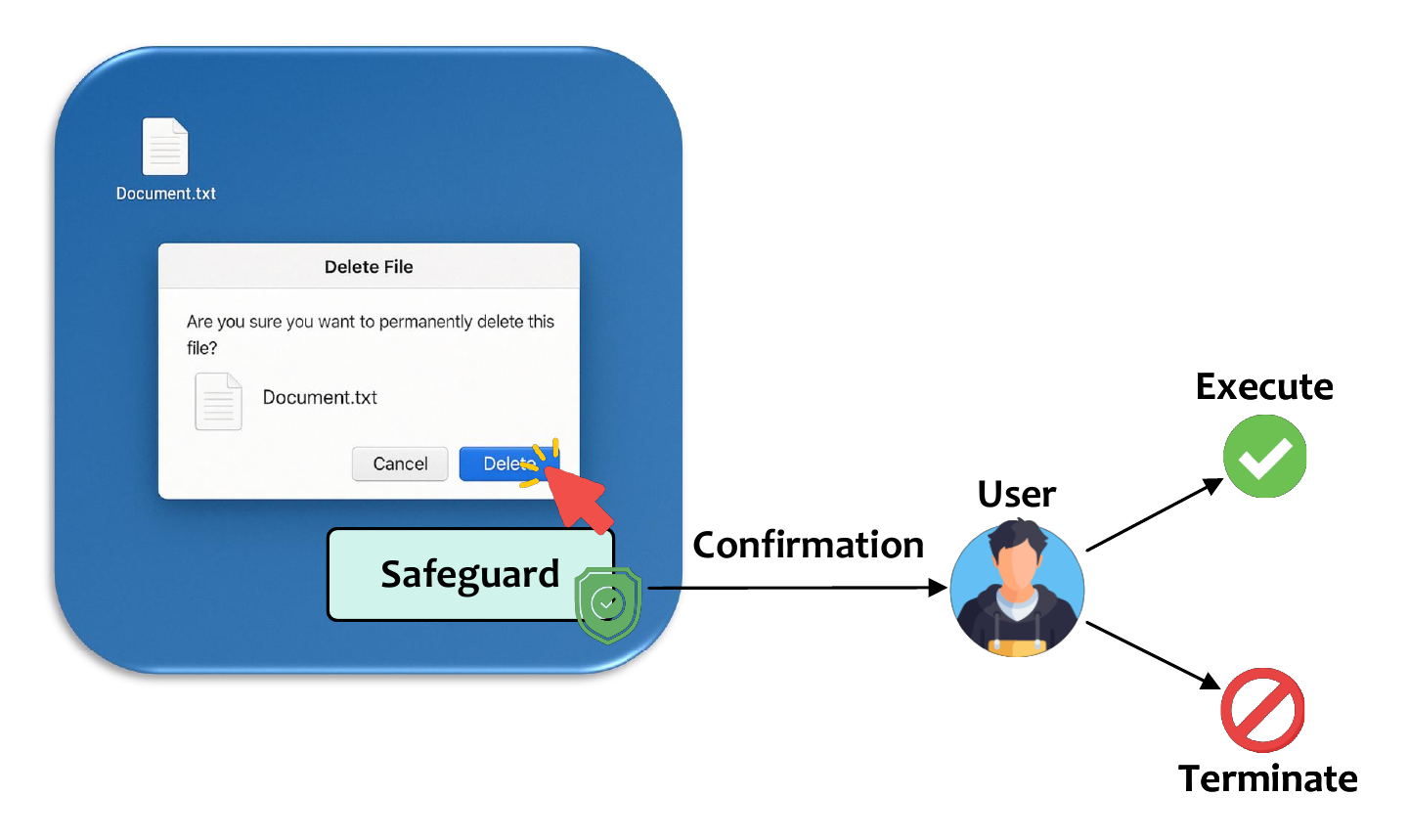}
    \vspace{-2em}
    \caption{The safeguard mechanism employed in \name.}
    \label{fig:safeguard}
    \vspace{-1em}
\end{figure}
While automation substantially boosts productivity, any CUA carries inherent risks of executing unsafe actions that may adversely affect user data or system stability~\cite{St-webagentbench, zhang2024ufo}. Examples include deleting critical files, terminating applications prematurely (resulting in unsaved data loss), or activating sensitive devices such as webcams without explicit consent. These actions pose severe risks, potentially causing irrecoverable damage or security breaches.

To mitigate such risks, \name incorporates an explicit \emph{safeguard mechanism}, designed to actively detect potentially dangerous actions, as shown in Figure~\ref{fig:safeguard}. Specifically, whenever an \appa identifies an action matching predefined risk criteria, it transitions into a dedicated \texttt{PENDING} state, pausing execution and actively prompting the user for confirmation. Only upon receiving explicit user consent does the agent proceed; otherwise, the action is aborted to prevent harm. The definition and scope of what constitutes a risky action are fully customizable through a straightforward prompt-based interface, enabling users and system administrators to precisely tailor safeguard behavior according to their organization's specific risk policies. This flexibility allows the safeguard system to be dynamically adapted as automation requirements evolve.

Through this proactive safety-checking framework, \name significantly reduces the likelihood of executing harmful operations, thus enhancing overall system safety, user trust, and robustness in real-world deployments.

\subsection{Everything-as-an-\appa}
\label{sec:everything}
\begin{figure}[t]
    \centering
    \includegraphics[width=\columnwidth]{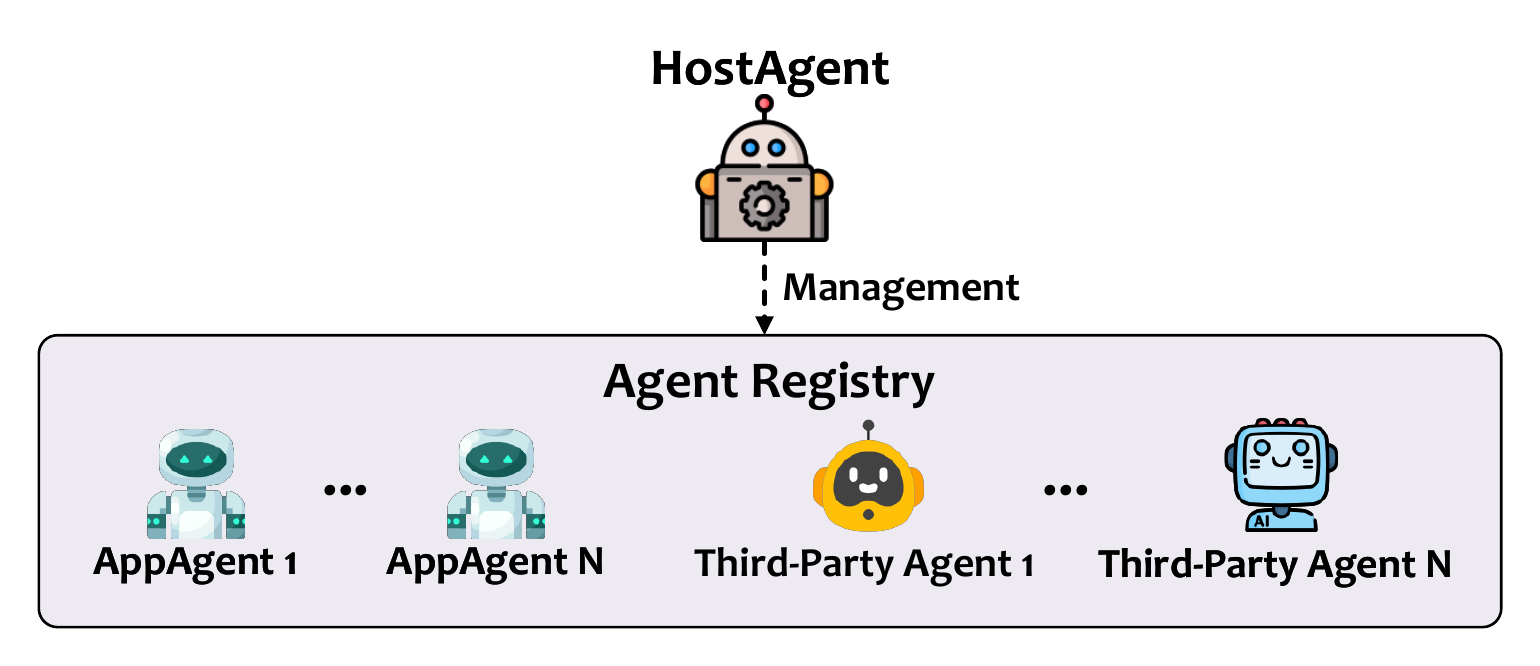}
    \vspace{-2em}
    \caption{The agent registry supports seamless wrapping of third-party components into the \appa framework.}
    \label{fig:everything}
    \vspace{-1em}
\end{figure}

To support ecosystem extensibility, \name introduces an agent registry mechanism that encapsulates arbitrary third-party components as pluggable \appas (Figure~\ref{fig:everything}). Through a simple registration API, external automation solutions—such as domain-specific copilots or proprietary tools—can be wrapped with lightweight compatibility shims that expose a unified interface to the \hosta.

This design enables \hosta to treat native and external \appas interchangeably, dispatching subtasks based on capability and specialization. We find that even minimal wrappers (\eg for OpenAI Operator~\cite{cua2025}) lead to tangible performance gains, highlighting the system's modularity and its ability to incorporate diverse execution backends with minimal engineering overhead.

\subsection{AgentOS-as-a-Service}
\label{sec:cloud}
\begin{figure}[t]
    \centering
    \includegraphics[width=\columnwidth]{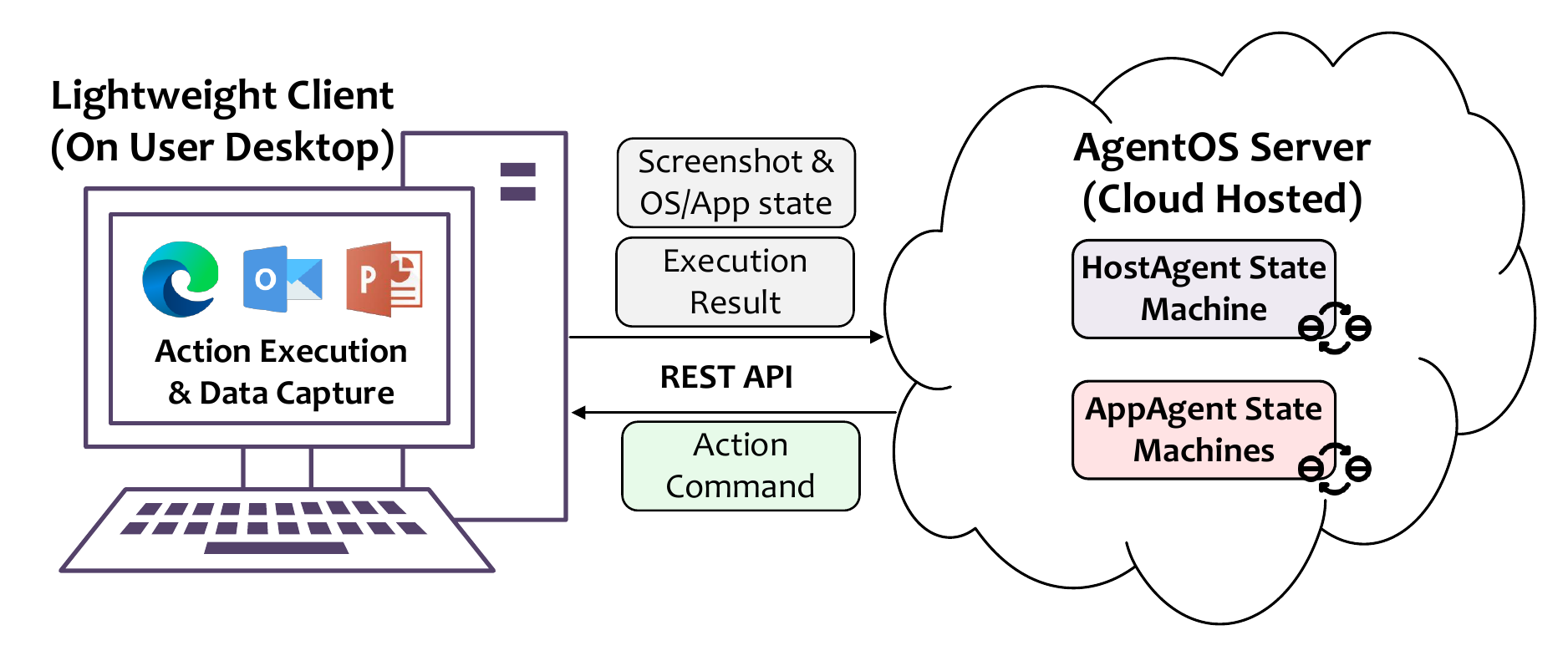}
    \vspace{-2em}
    \caption{The client-server deployment model used in AgentOS-as-a-Service.}
    \label{fig:server_client}
    \vspace{-1em}
\end{figure}

\name adopts a client-server architecture to support practical deployment at scale (Figure~\ref{fig:server_client}). A lightweight client resides on the user's machine and is responsible for GUI operations and application-side sensing. Meanwhile, a centralized server (running on-premises or in the cloud) hosts the \hosta/\appa logic, orchestrates workflows, and handles LLM queries.

This separation of control and execution offers several systems-level benefits:
\begin{itemize}
    \item \textbf{Security}: Sensitive orchestration and model execution are isolated from user devices.
    \item \textbf{Maintainability}: Server-side updates propagate without modifying the client.
    \item \textbf{Scalability}: The system can support multiple concurrent clients with centralized scheduling and load management.
\end{itemize}
The client-server boundary enforces a clean service abstraction, promoting modularity and simplifying rollout in enterprise environments.

\subsection{Comprehensive Logging and Debugging Infrastructure}
\label{sec:logging}

\begin{figure*}[t]
    \centering
    \includegraphics[width=0.8\textwidth]{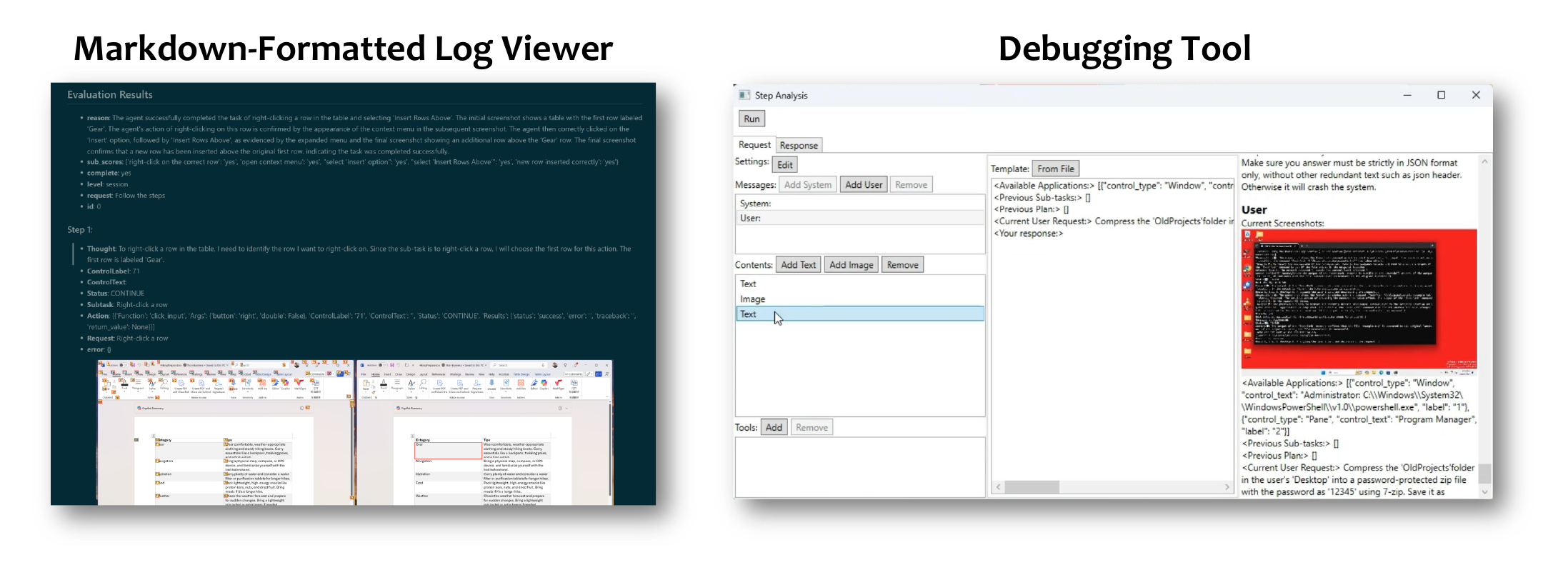}
    \vspace{-2.em}
    \caption{An illustration of the markdown-formatted log viewer and debugging tool in \name.}
    \label{fig:debug}
\end{figure*}

Robust observability is essential for diagnosing failures and supporting ongoing system improvement. To this end, \name implements a comprehensive logging and debugging framework. Each session captures fine-grained traces of execution: prompts, LLM outputs, control metadata, UI state snapshots, and error events.

At the end of each session, \name compiles these artifacts into a structured, Markdown-formatted execution log. Developers can inspect action-by-action agent decisions, visualize interface state transitions, and replay behavior for debugging. The framework also supports prompt editing and selective replay for targeted hypothesis testing, significantly accelerating the debugging cycle. We show an example of these tools in Figure~\ref{fig:debug}.

This observability layer functions as a lightweight provenance system for agent behavior, fostering transparency, accountability, and rapid iteration during deployment.

\subsection{Automated Task Evaluator}
\label{sec:evaluator}
\begin{figure}[t]
    \centering
    \includegraphics[width=\columnwidth]{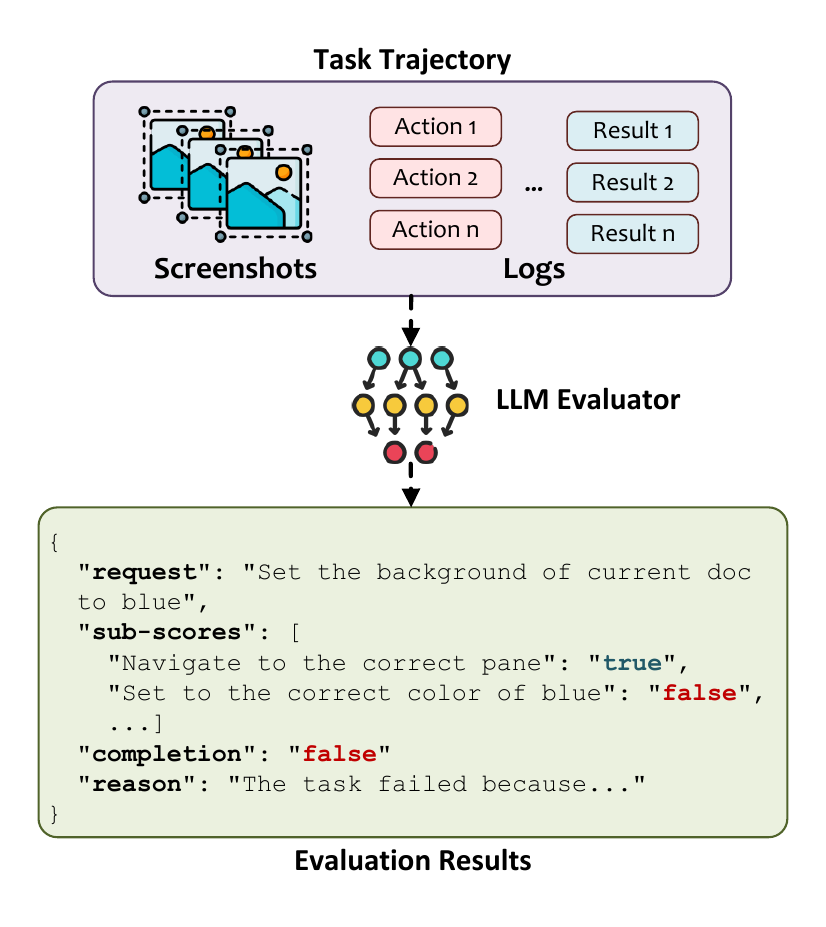}
    \vspace{-3.5em}
    \caption{The LLM-based task evaluator applies CoT reasoning to structured session logs.}
    \label{fig:evaluator}
    \vspace{-1em}
\end{figure}

To provide structured feedback and facilitate continuous improvement, \name includes an automated task evaluation engine based on LLM-as-a-judge~\cite{chen2024mllm}. As shown in Figure~\ref{fig:evaluator}, the evaluator parses session traces—including actions, rationales, and screenshots—and applies CoT reasoning to decompose tasks into evaluation criteria.

It assigns partial scores and synthesizes an overall result: \emph{success}, \emph{partial}, or \emph{failure}. This structured outcome feeds into downstream dashboards and debugging tools. It also supports self-monitoring and offline analysis of failure cases, closing the loop between execution, diagnosis, and improvement.

\vspace{0.5em}
\noindent
\textbf{Summary.} These engineering components demonstrate \name's commitment to operational robustness and extensibility. From session-based execution and pluggable agents to service-oriented deployment and observability infrastructure, each module reflects a design focused on bridging conceptual LLM agent architectures with the systems realities of deployment at scale.

\section{Evaluation} 
We tested \name rigorously across more than 20 Windows applications, including office suites, file explorers, and custom enterprise tools to assess performance, efficiency, and robustness. Our experiments show:
\begin{enumerate} 
    \item \name achieves a \textit{10\%} higher task completion rate, a 50\% relative improvement—over the best-performing current CUA Operator, enabled by deeper OS-level integration.
    \item The hybrid UIA--vision approach identifies custom or nonstandard GUI elements missed by UIA alone, boosting success in interfaces with proprietary widgets. 
    \item Allowing \appas to invoke native APIs or GUI interactions to improve completion rate by over 8\%, cuts latency and reduces the fragility seen in purely click-based workflows. 
    \item Leveraging external documents and execution logs increases \name's ability to handle unfamiliar features without retraining. 
    \item Speculative multi-action execution consolidates multiple steps into a single LLM call, lowering inference cost by up to \textit{51.5\%} without compromising reliability. 
    \item  By enabling Everything-as-an-\appa (\eg Operator), \name both boosts overall performance and uncovers the full potential of each individual agent.
\end{enumerate}
Overall, these results confirm that \name's deeper integration with Windows and application-level APIs yields both higher performance and reduced overhead, making a compelling case for an OS-native approach to desktop automation.

\subsection{Experimental Setup}

\paragraph{Deployment Environment.} 
The benchmark environments are hosted on isolated VMs with \textit{8} AMD Ryzen 7 CPU cores and \textit{8} GB of memory, matching typical deployment conditions.  All GPT-family models (GPT-4V, GPT-4o, o1, and Operator) are accessed via Azure OpenAI services, while the OmniParser-v2 vision model operates on a separate virtual machine provisioned with an NVIDIA A100 80GB GPU to support efficient and high-throughput visual grounding.

\paragraph{Benchmarks.} 
We evaluate \name using two established Windows-centric automation benchmarks:
\begin{itemize} 
    \item \textbf{Windows Agent Arena (WAA)~\cite{bonatti2024windows}:} 
    Consists of 154 live automation tasks across 15 commonly used Windows applications, including office productivity tools, web browsers, system utilities, development environments, and multimedia apps. Each task includes a custom verification script for automated correctness checking.

    \item \textbf{OSWorld-W~\cite{xie2024osworld}:} 
    A targeted subset of the OSWorld benchmark specifically tailored for Windows, comprising 49 live tasks across office applications, browser interactions, and file-system operations. Tasks are similarly equipped with handcrafted verification scripts for reliable outcome validation.
\end{itemize}

Each task runs independently, and verification strictly follows the original scripts provided by each benchmark.\footnote{Reported baseline scores in OSWorld differ slightly from prior results focused on Ubuntu due to corrections in verification scripts and alignment with Windows-specific tasks (OSWorld-W).}

\paragraph{Baselines.} 
We compare \name with five representative state-of-the-art CUAs, each leveraging GPT-4o as the inference engine:

\begin{itemize} 
    \item \textbf{UFO}~\cite{zhang2024ufo}: A pioneering multiagent, GUI-focused automation system designed explicitly for Windows, integrating UIA and visual perception.
    \item \textbf{NAVI}~\cite{bonatti2024windows}: A single-agent baseline from WAA, utilizing screenshots and accessibility data for GUI understanding.
    \item \textbf{OmniAgent}~\cite{lu2024omniparser}: Employs OmniParser for visual grounding combined with GPT-based action planning.
    \item \textbf{Agent S}~\cite{agasheagent}: Features a multiagent architecture with experience-driven hierarchical planning, optimized for complex, multi-step tasks.
    \item \textbf{Operator}~\cite{cua2025}: A recent, high-performance CUA from OpenAI, simulating human-like mouse and keyboard interactions via screenshots.
\end{itemize}
These baselines were selected for their representativeness of diverse architectural and design paradigms (\eg single-agent vs. multiagent, GUI-only vs. hybrid approaches). To ensure fairness, each agent is restricted to a maximum of \textit{30 execution steps} per task, reflecting practical user expectations and preventing excessively long task executions. Additionally, we evaluate a base version of \name (termed \name-base) using only UIA detection, GUI-based interactions, and without dynamic knowledge integration, alongside the full implementation of \name featuring hybrid control detection, combined GUI-API interactions, and continuous knowledge augmentation. API integrations were selectively implemented for three office applications within OSWorld-W as illustrative examples; no APIs were introduced for the WAA tasks. Further implementation details are available in Section~\ref{sec:eva_api}.

\paragraph{Evaluation Metrics.} 
We utilize two primary metrics for performance evaluation:

\begin{itemize} 
    \item \textbf{Success Rate (SR):} Defined as the percentage of tasks successfully completed, validated via the benchmarks' own verification scripts.
    \item \textbf{Average Completion Steps (ACS):} Measures the average number of LLM-involved action inference steps required per task. Fewer steps correspond to higher efficiency, directly correlating with lower inference latency and reduced computational overhead.
\end{itemize}
These metrics effectively reflect both functional effectiveness and practical efficiency, providing clear indicators of real-world automation performance.

\subsection{Success Rate Comparison}

\begin{table}[t]
\centering
\caption{Comparison of success rates (SR) across agents on WAA and OSWorld-W benchmarks.}
\label{tab:compare}
\begin{tabular}{l|l|c|c}
\hline
\textbf{Agent} & \textbf{Model} & \textbf{WAA} & \textbf{OSWorld-W} \\ \hline
UFO            & GPT-4o         & 19.5\%       & 12.2\%                   \\ \hline
NAVI           & GPT-4o         & 13.3\%       & 10.2\%                   \\ \hline
OmniAgent      & GPT-4o         & 19.5\%       & 8.2\%                   \\ \hline
Agent S        & GPT-4o         & 18.2\%       & 12.2\%                   \\ \hline
Operator       & computer-use   & 20.8\%       & 14.3\%             \\ \hline\hline
\name-base     & GPT-4o         & 23.4\%        & 16.3\%                    \\ \hline
\name-base     & o1             & 25.3\%        & 16.3\%                    \\ \hline
\textbf{\name} & \textbf{GPT-4o}& \textbf{27.9\%}&  \textbf{28.6\%}         \\ \hline
\textbf{\name} & \textbf{o1}    & \textbf{30.5\%}&   \textbf{32.7\%}        \\ \hline\hline
\end{tabular}
\end{table}

Table~\ref{tab:compare} summarizes the success rates (SR) of all evaluated agents across the WAA and OSWorld-W benchmarks, as verified by each benchmark's automated validation scripts. Notably, even the basic configuration (\name-base)—which relies solely on standard UI Automation and GUI-driven actions—consistently surpasses prior state-of-the-art CUAs. Specifically, with GPT-4o, \name-base achieves an SR of 23.4\% on WAA, outperforming the best existing baseline, Operator (20.8\%), by 2.6\%. This margin widens significantly when employing the stronger o1 LLM, lifting \name-base's performance to 25.3\%.

Moreover, the complete version of \name, incorporating hybrid GUI–API action execution, advanced visual grounding, and continuous knowledge integration, further amplifies these performance gains. With GPT-4o, \name achieves a 27.9\% SR on WAA, exceeding Operator by a substantial 7.1\%. The performance gap becomes even more pronounced on OSWorld-W, where \name achieves a 28.6\% SR compared to Operator's 14.3\%, effectively doubling its success rate. Utilizing the stronger o1 model further improves \name's performance to 30.5\% (WAA) and 32.7\% (OSWorld-W), solidifying its leading position.

These significant performance improvements clearly underscore the advantages of \name's deep integration with OS-level mechanisms and its unified system architecture. While prior CUAs primarily emphasize model-level optimization or singular reliance on visual interfaces, our results demonstrate that robust, system-level orchestration—combining structured OS APIs, specialized application knowledge, and hybrid GUI–API interaction—is instrumental in achieving higher task reliability and broader automation coverage. Crucially, even a general-purpose, less-specialized model like GPT-4o can surpass highly specialized CUAs (such as Operator) when integrated within the comprehensive \name framework. This insight reinforces the value of architectural design and OS integration as key drivers of practical, deployable desktop automation solutions.

\begin{table*}[t]
\caption{SR breakdown by application type on WAA and OSWorld-W. \label{tab:breakdown}}
\begin{tabular}{l|l|cccccc|cc}
\hline
\multirow{2}{*}{\textbf{Agent}} & \multirow{2}{*}{\textbf{Model}} & \multicolumn{6}{c|}{\textbf{WAA}}                                                                                                                                                                                                                                                                                                                                                   & \multicolumn{2}{c}{\textbf{OSWorld-W}}                \\ \cline{3-10} 
                                &                                 & \multicolumn{1}{c|}{Office}       & \multicolumn{1}{c|}{\begin{tabular}[c]{@{}c@{}}Web \\ Browser\end{tabular}} & \multicolumn{1}{c|}{\begin{tabular}[c]{@{}c@{}}Windows \\ System\end{tabular}} & \multicolumn{1}{c|}{Coding}          & \multicolumn{1}{c|}{\begin{tabular}[c]{@{}c@{}}Media \& \\ Video\end{tabular}} & \begin{tabular}[c]{@{}c@{}}Windows \\ Utils\end{tabular} & \multicolumn{1}{c|}{Office}          & Cross-App      \\ \hline
UFO                             & GPT-4o                          & \multicolumn{1}{c|}{0.0\%}        & \multicolumn{1}{c|}{23.3\%}                                                 & \multicolumn{1}{c|}{33.3\%}                                                    & \multicolumn{1}{c|}{29.2\%}          & \multicolumn{1}{c|}{33.3\%}                                                    & 8.3\%                                                    & \multicolumn{1}{c|}{18.5\%}          & 4.5\%          \\ \hline
NAVI                            & GPT-4o                          & \multicolumn{1}{c|}{0.0\%}        & \multicolumn{1}{c|}{20.0\%}                                                 & \multicolumn{1}{c|}{29.2\%}                                                    & \multicolumn{1}{c|}{9.1\%}           & \multicolumn{1}{c|}{25.3\%}                                                    & 0.0\%                                                    & \multicolumn{1}{c|}{18.5\%}          & 0.0\%          \\ \hline
OmniAgent                       & GPT-4o                          & \multicolumn{1}{c|}{0.0\%}        & \multicolumn{1}{c|}{27.3\%}                                                 & \multicolumn{1}{c|}{33.3\%}                                                    & \multicolumn{1}{c|}{27.3\%}          & \multicolumn{1}{c|}{30.3\%}                                                    & 8.3\%                                                    & \multicolumn{1}{c|}{14.8\%}          & 0.0\%          \\ \hline
Agent S                         & GPT-4o                          & \multicolumn{1}{c|}{0.0\%}        & \multicolumn{1}{c|}{13.3\%}                                                 & \multicolumn{1}{c|}{45.8\%}                                                    & \multicolumn{1}{c|}{29.2\%}          & \multicolumn{1}{c|}{19.1\%}                                                    & \textbf{22.2\%}                                          & \multicolumn{1}{c|}{22.2\%}          & 0.0\%          \\ \hline
Operator                        & computer-use                    & \multicolumn{1}{c|}{\textbf{7\%}} & \multicolumn{1}{c|}{26.7\%}                                                 & \multicolumn{1}{c|}{29.2\%}                                                    & \multicolumn{1}{c|}{29.2\%}          & \multicolumn{1}{c|}{28.6\%}                                                    & 8.3\%                                                    & \multicolumn{1}{c|}{22.2\%}          & 4.5\%          \\ \hline\hline
\name-base                      & GPT-4o                          & \multicolumn{1}{c|}{2.3\%}        & \multicolumn{1}{c|}{36.7\%}                                                 & \multicolumn{1}{c|}{29.2\%}                                                    & \multicolumn{1}{c|}{41.7\%}          & \multicolumn{1}{c|}{33.3\%}                                                    & 0.0\%                                                    & \multicolumn{1}{c|}{22.2\%}          & \textbf{9.1\%} \\ \hline
\name-base                      & o1                              & \multicolumn{1}{c|}{2.3\%}        & \multicolumn{1}{c|}{30.0\%}                                                 & \multicolumn{1}{c|}{37.5\%}                                                    & \multicolumn{1}{c|}{50.0\%}          & \multicolumn{1}{c|}{33.3\%}                                                    & 8.3\%                                                    & \multicolumn{1}{c|}{22.2\%}          & \textbf{9.1\%} \\ \hline
\name                           & GPT-4o                          & \multicolumn{1}{c|}{4.7\%}        & \multicolumn{1}{c|}{30.0\%}                                                 & \multicolumn{1}{c|}{41.7\%}                                                    & \multicolumn{1}{c|}{\textbf{58.3\%}} & \multicolumn{1}{c|}{33.3\%}                                                    & 8.3\%                                                    & \multicolumn{1}{c|}{44.4\%}          & \textbf{9.1\%} \\ \hline
\name                           & o1                              & \multicolumn{1}{c|}{4.7\%}        & \multicolumn{1}{c|}{\textbf{40.0\%}}                                        & \multicolumn{1}{c|}{\textbf{45.8\%}}                                           & \multicolumn{1}{c|}{50.0\%}          & \multicolumn{1}{c|}{\textbf{38.1\%}}                                           & 16.7\%                                                   & \multicolumn{1}{c|}{\textbf{51.9\%}} & \textbf{9.1\%} \\ \hline
\end{tabular}
\end{table*}

\paragraph{Performance Breakdown.} 
Table~\ref{tab:breakdown} presents a detailed breakdown of success rates (SR) by application type on the WAA and OSWorld-W benchmarks, enabling deeper understanding of where \name achieves particularly strong results and identifying areas for further system-level improvements. Across multiple categories, \name consistently demonstrates superior performance compared to baseline CUAs, particularly in application scenarios demanding deeper OS integration or sophisticated multi-step task execution.

Notably, \name excels in tasks involving web browsers and coding environments. For instance, the strongest configuration (\name with o1) attains an impressive 40.0\% SR for web browser tasks—markedly outperforming the next-best baseline (OmniAgent) by over 12\%. Similarly, in coding-related workflows, \name (GPT-4o) achieves the highest SR of 58.3\%, significantly exceeding all competing CUAs. These results underscore the effectiveness of \name's hybrid GUI-API approach and continuous knowledge integration, which enable more precise action inference, reduce brittleness due to GUI changes, and substantially elevate reliability in multi-step workflows.

The breakdown further reveals a clear correlation between application complexity, popularity, and system-level support. Tasks involving LibreOffice (in the Office category of WAA) uniformly yield lower SRs across all evaluated CUAs, largely due to inadequate adherence to accessibility standards and incomplete UIA support. Conversely, OSWorld-W tasks predominantly utilize Microsoft 365 Office applications, which offer richer OS-native APIs and structured accessibility data, resulting in improved SRs (up to 51.9\% for \name-o1). This discrepancy highlights the critical role that robust OS-level integration and API availability play in achieving high-quality desktop automation.

Cross-application tasks, especially prominent in OSWorld-W, present an even greater challenge. Such tasks inherently require sophisticated task decomposition and robust inter-agent coordination, pushing CUAs—and even human users—to their limits. Here, \name's multiagent architecture, led by the centralized \hosta and specialized \appas, demonstrates notable promise, outperforming other baselines with a 9.1\% SR. Although performance remains relatively modest, it clearly illustrates the strength of systematic multiagent collaboration and centralized orchestration in addressing complex scenarios that cross traditional application boundaries.

Overall, these detailed breakdown results validate the system-level design principles of \name, particularly its emphasis on deep OS and application-specific integration, multiagent coordination, and flexible action orchestration. While there remains significant potential for further enhancements in niche or less-supported application domains (\eg custom or legacy software with limited API availability), \name's current architecture already provides a substantial, measurable improvement in practical, real-world desktop automation tasks.

\begin{figure}[t]
    \centering
    \includegraphics[width=\columnwidth]{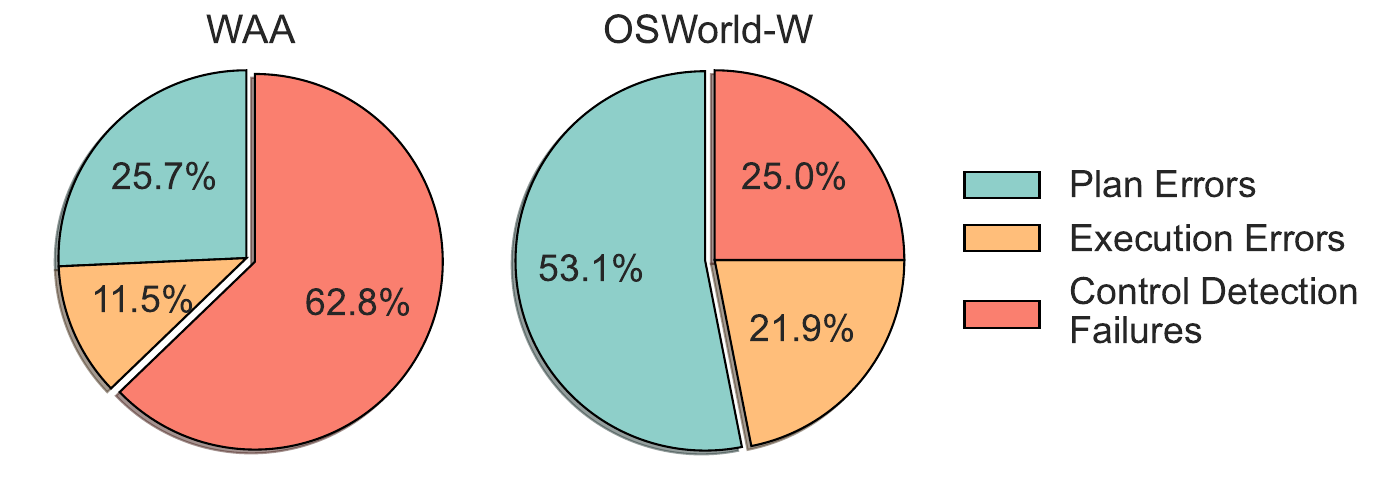}
    \vspace{-2em}
    \caption{Error analysis of \name-base (GPT-4o) on the two benchmarks.}
    \label{fig:error_ana}
    \vspace{-1em}
\end{figure}

\paragraph{Error Analysis.}
To systematically understand the limitations of \name and identify opportunities for further improvement, we conducted a detailed manual review of all failure cases for \name-base (GPT-4o) on both benchmarks. Following a classification framework similar to Agashe~\etal~\cite{agasheagent}, each failure was categorized into one of three distinct system-level categories:
\begin{itemize} 
    \item \textbf{Plan Errors:} Failures arising from inadequate high-level task understanding, typically reflected by incomplete or incorrect action plans. These errors indicate gaps in the agent's task comprehension or insufficient grounding in application-specific workflows.
    
    \item \textbf{Execution Errors:} Cases where the high-level plan is reasonable but the execution is flawed (\eg selecting an incorrect control, performing unintended actions). Execution errors often stem from inaccurate visual reasoning, incorrect associations between GUI elements and actions, or erroneous inference by the LLM.
    
    \item \textbf{Control Detection Failures:} Instances where the agent fails to detect or identify critical GUI controls required to complete a task, usually due to non-standard or custom-rendered UI elements that are not fully accessible via standard OS APIs.
\end{itemize}
Figure~\ref{fig:error_ana} summarizes our findings for \name-base. On the WAA benchmark, more than 62\% of failures were attributed to \emph{Control Detection Failures}, highlighting significant gaps in standard UIA API coverage—especially for third-party applications (\eg LibreOffice) that do not strictly adhere to accessibility standards. Conversely, the OSWorld-W benchmark exhibited a higher incidence of \emph{Plan Errors}, underscoring that tasks in this set frequently involve more complex workflows, necessitating deeper domain knowledge or advanced contextual reasoning capabilities beyond simple visual recognition.

These observations provide concrete evidence of specific system-level shortcomings, directly motivating the enhancements incorporated into the complete version of \name. The high frequency of \emph{Control Detection Failures} validates our choice of adopting a hybrid GUI detection pipeline that supplements standard UIA data with advanced visual grounding techniques. Similarly, the prevalence of \emph{Plan Errors} underscores the critical role of integrating richer external documentation, domain-specific knowledge bases, and application-level APIs to strengthen task understanding and action inference. In the subsequent sections, we explicitly demonstrate how these incremental system-level improvements progressively mitigate each identified category of errors, thereby substantially boosting \name's overall task completion effectiveness.


\subsection{Evaluation on Hybrid Control Detection}
\begin{table}[t]
\caption{Comparison of SR and CRR across control detection mechanisms.}
\label{tab:control}
\resizebox{\columnwidth}{!}{ 
\begin{tabular}{l|l|cc|cc}
\hline
\multirow{2}{*}{\textbf{Control Detector}} & \multirow{2}{*}{\textbf{Model}} & \multicolumn{2}{c|}{\textbf{WAA}}                     & \multicolumn{2}{c}{\textbf{OSWorld-W}}                 \\ \cline{3-6} 
                                           &                                 & \multicolumn{1}{c|}{SR}              & CRR            & \multicolumn{1}{c|}{SR}              & CRR             \\ \hline
UIA                                        & GPT-4o                          & \multicolumn{1}{c|}{23.4\%}          & -              & \multicolumn{1}{c|}{22.4\%}          & -               \\ \hline
OmniParser-v2                              & GPT-4o                          & \multicolumn{1}{c|}{26.6\%}          & 7.0\%          & \multicolumn{1}{c|}{14.3\%}          & 0\%             \\ \hline
\textbf{Hybrid}                            & \textbf{GPT-4o}                 & \multicolumn{1}{c|}{\textbf{26.6\%}} & \textbf{9.9\%} & \multicolumn{1}{c|}{\textbf{22.4\%}} & \textbf{12.5\%} \\ \hline\hline
UIA                                        & o1                              & \multicolumn{1}{c|}{25.3\%}          & -              & \multicolumn{1}{c|}{24.5\%}          & -               \\ \hline
OmniParser-v2                              & o1                              & \multicolumn{1}{c|}{20.8\%}          & 7.0\%          & \multicolumn{1}{c|}{14.3\%}          & 0\%             \\ \hline
\textbf{Hybrid}                            & \textbf{o1}                     & \multicolumn{1}{c|}{\textbf{27.9\%}} & \textbf{9.9\%} & \multicolumn{1}{c|}{\textbf{28.6\%}} & \textbf{25.0\%} \\ \hline
\end{tabular}
}
\end{table}


\begin{figure}[t]
    \centering
    \includegraphics[width=\columnwidth]{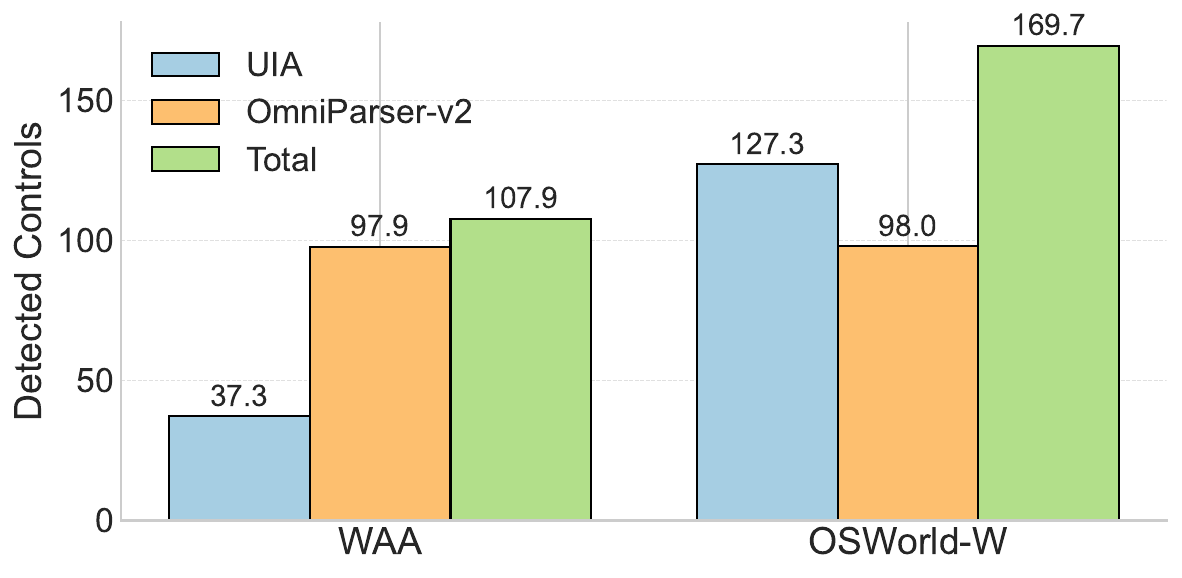}
    \vspace{-2em}
    \caption{The number of detected controls of different approaches.}
    \label{fig:control_num}
    \vspace{-1em}
\end{figure}

As shown in Figure~\ref{fig:error_ana}, a considerable fraction of failures arise from \emph{Control Detection Failures}, where non-standard UI elements do not comply with UIA guidelines. To quantify the effectiveness of different detection strategies, we compare UIA-only, OmniParser-v2--only, and our hybrid method (Section~\ref{sec:hybrid-control}). We introduce a \emph{Control Recovery Ratio (CRR)} to measure how many UIA-only failures are ``recovered'' (\ie become successful completions) under OmniParser or the hybrid approach.

Table~\ref{tab:control} presents the results on both benchmarks, across multiple model configurations. The hybrid method consistently outperforms either UIA-only or OmniParser-only settings, raising the overall success rate and converting up to 9.86\% of previously irrecoverable cases into completions. This gain highlights the complementary strengths of the two detection pipelines, as the hybrid approach bridges coverage gaps in UIA while avoiding OmniParser's limitations in more standardized GUIs.

In Figure~\ref{fig:control_num}, we report the average number of controls detected from each source (UIA, OmniParser-v2, and the merged set) under the hybrid approach. Owing to differences in application coverage, the total number of detected controls is generally higher in OSWorld-W than in WAA. Notably, both UIA and OmniParser-v2 identify substantial subsets of controls, and after merging, 27.9\% and 56.7\% of OmniParser-v2 detections are discarded due to overlap with UIA. These observations indicate that OmniParser-v2 provides a valuable complement to UIA by recovering nonstandard or custom elements. At the same time, the merging step removes redundancies and prevents double-counting, ultimately reducing control detection failures in the hybrid scheme.

\subsection{Effectiveness of GUI + API Integration \label{sec:eva_api}}
\begin{table}[t]
\centering
\caption{APIs supported across Office applications.}
\label{tab:office_apis}
\resizebox{\columnwidth}{!}{ 
\begin{tabular}{@{}l|c|p{4cm}@{}}
\toprule
\textbf{API} & \textbf{Application} & \textbf{Description} \\ \midrule
\texttt{select\_text}          & Word       & Select matched text in the document. \\\hline
\texttt{select\_paragraph}     & Word       & Select a paragraph in the document. \\\hline
\texttt{set\_font}             & Word       & Set the font size and style of selected text. \\\hline
\texttt{save\_as}              & Word       & Save the current document to a desired format. \\\hline
\texttt{insert\_excel\_table}  & Excel      & Insert a table at the desired position. \\\hline
\texttt{select\_table\_range}  & Excel      & Select a range within a table. \\\hline
\texttt{reorder\_column}       & Excel      & Reorder columns of a table. \\\hline
\texttt{save\_as}              & Excel      & Save the current sheet to a desired format. \\\hline
\texttt{set\_background\_color}& PowerPoint & Set the background color of slide(s). \\\hline
\texttt{save\_as}              & PowerPoint & Save the current presentation to a desired format. \\
\bottomrule
\end{tabular}
}
\end{table}

\begin{table}[t]
\caption{Performance comparison of GUI-only vs. GUI + API actions.}
\label{tab:api_compare}
\resizebox{\columnwidth}{!}{ 
\begin{tabular}{l|l|c|c|c|c|c}
\hline
\textbf{Action}  & \textbf{Model}  & \textbf{SR}     & \textbf{PRR}    & \textbf{ERR}    & \textbf{CRR}    & \textbf{ACS}  \\ \hline
GUI-only         & GPT-4o          & 16.3\%          & -               & -               & -               & 13.8          \\ \hline
\textbf{GUI+API} & \textbf{GPT-4o} & \textbf{22.4\%} & \textbf{5.9\%}  & \textbf{14.3\%} & \textbf{25.0\%} & \textbf{12.9} \\ \hline\hline
GUI-only         & o1              & 16.3\%          & -               & -               & -               & 16.0          \\ \hline
\textbf{GUI+API} & \textbf{o1}     & \textbf{24.5\%} & \textbf{17.7\%} & \textbf{0.0\%}  & \textbf{12.5\%} & \textbf{6.6}  \\ \hline
\end{tabular}
}
\end{table}

\begin{figure*}[t]
    \centering
    \includegraphics[width=\textwidth]{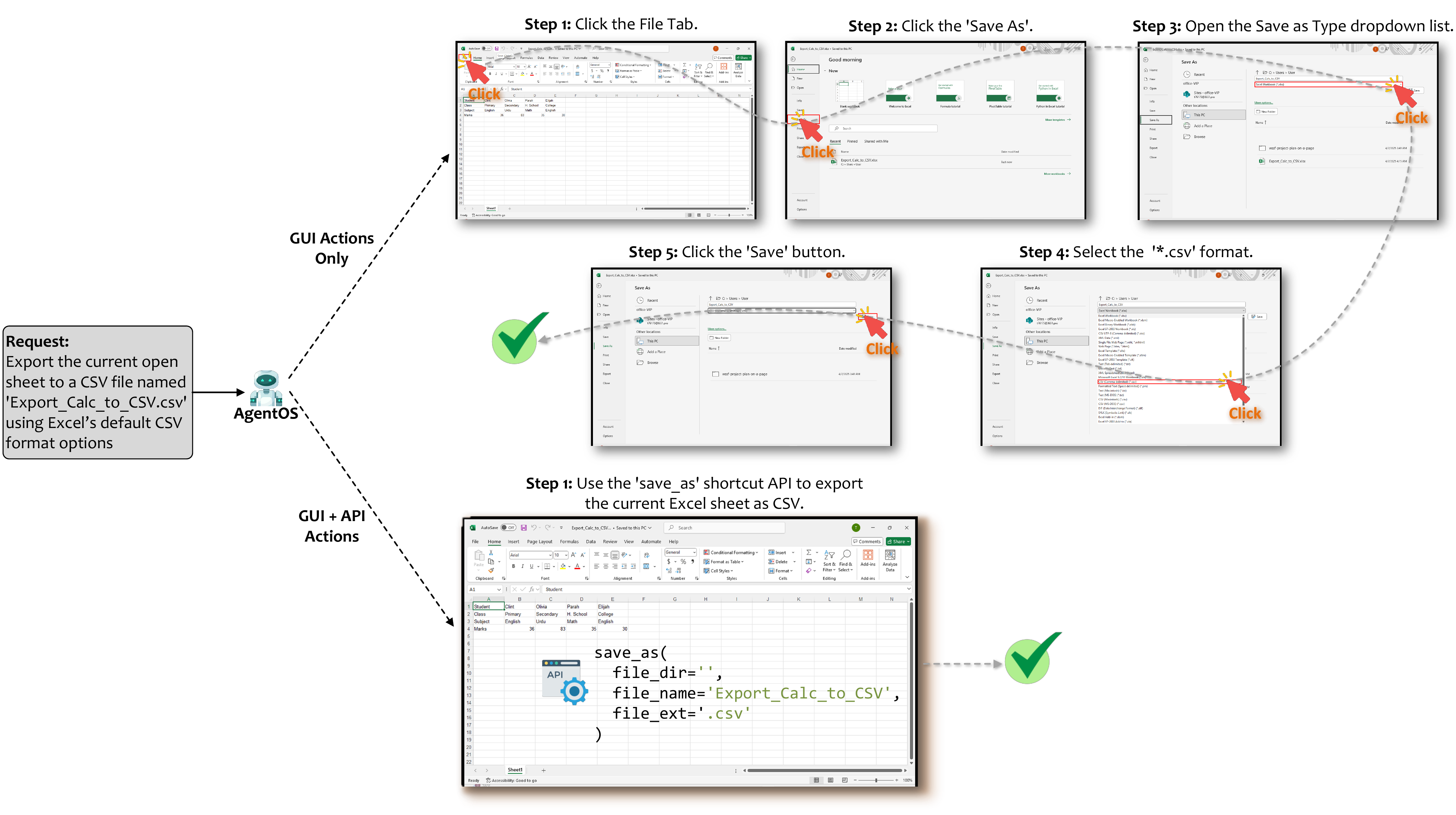}
    \vspace{-2em}
    \caption{A case study comparing the completion of the same task using GUI-only actions vs. GUI + API actions.}
    \label{fig:api_gui_case}
\end{figure*}

We now evaluate how unifying API-based actions with standard GUI interactions in the \texttt{Puppeteer} impacts performance (Section~\ref{sec:api}). To do so, we focus on the \textit{27} office-related tasks in OSWorld and manually develop \textit{12} APIs for Word, Excel, and PowerPoint. These applications provide COM interfaces that facilitate the creation of custom functions, making them ideal exemplars for deeper OS- and application-level integration. Importantly, many of these operations would require cumbersome multi-step GUI procedures but become straightforward single calls via these APIs (\eg select paragraphs). Table~\ref{tab:office_apis} details the implemented APIs.

Table~\ref{tab:api_compare} compares \textit{(i)} overall Success Rate (SR), \textit{(ii)} Plan Error Recovery rate (PRR), \textit{(iii)} Execution Error Recovery rate (ERR), \textit{(iv)} Control Detection Failure Recovery rate (CRR), and \textit{(v)} Average Completion Steps (ACS) for two configurations: GUI-only versus GUI + API. We calculate ACS on the subset of tasks that both configurations successfully complete, ensuring a fair comparison.

The results show that integrating API actions boosts SR for both GPT-4o (+6.1\%) and o1 (+8.2\%), underscoring the effectiveness of mixing GUI and API interactions. Notably, GPT-4o benefits most from APIs in recovering from \emph{Control Detection Failures} by circumventing unannotated GUI elements. In contrast, o1 more frequently addresses \emph{Plan Errors} through API ``shortcuts'', reflecting the model's stronger reasoning capabilities and preference for concise solutions.

Beyond higher success rates, GUI + API also reduces the effort required to complete tasks. \name achieves a 6.5\% step savings with GPT-4o and an impressive 58.5\% reduction for o1 on identical tasks. The latter improvement stems from o1's ability to strategically call API functions, bypassing multiple GUI-based steps. Overall, these findings confirm the advantages of mixing GUI automation with API calls, both in terms of robustness and efficiency, and showcase the importance of deep system integration for desktop automation.

\paragraph{Case Study.} To illustrate how the GUI + API approach streamlines task execution, Figure~\ref{fig:api_gui_case} shows the completion trajectory for exporting an Excel file to CSV format in a case of OSWorld-W, using either GUI-only or GUI + API interactions. Although both configurations eventually succeed, the GUI-only setting requires five steps to open the Save dialog, select the file format, and confirm the action. In contrast, a single call to the \texttt{save\_as} API completes the task immediately. Beyond improving efficiency, this one-step solution also reduces the risk of compounding errors across multiple GUI interactions—a clear demonstration of the advantages of deeper OS and application-level integration.

\subsection{Continuous Knowledge Integration Evaluation} \label{sec:knowledge_enhancement}

\begin{table}[t]
\caption{Performance comparison with and without knowledge integration.}
\label{tab:knowledge}
\resizebox{\columnwidth}{!}{ 
\begin{tabular}{l|l|cc|cc}
\hline
\multirow{2}{*}{\textbf{\begin{tabular}[c]{@{}l@{}}Knowledge\\ Enhancement\end{tabular}}} & \multirow{2}{*}{\textbf{Model}} & \multicolumn{2}{c|}{\textbf{WAA}}    & \multicolumn{2}{c}{\textbf{OSWorld-W}} \\ \cline{3-6} 
                                                                                          &                                 & \multicolumn{1}{c|}{SR}     & PRR    & \multicolumn{1}{c|}{SR}      & PRR     \\ \hline
None                                                                                      & GPT-4o                          & \multicolumn{1}{c|}{23.4\%} & -      & \multicolumn{1}{c|}{22.4\%}        & -        \\ \hline
Help Document                                                                             & GPT-4o                          & \multicolumn{1}{c|}{26.6\%} & 10.34\% & \multicolumn{1}{c|}{26.5\%}        & 11.8\%        \\ \hline
Self-Experience                                                                           & GPT-4o                          & \multicolumn{1}{c|}{26.6\%} & 13.79\% & \multicolumn{1}{c|}{24.5\%}        & 11.8\%        \\ \hline\hline
None                                                                                      & o1                              & \multicolumn{1}{c|}{25.3\%} & -      & \multicolumn{1}{c|}{24.5\%}        & -        \\ \hline
Help Document                                                                             & o1                              & \multicolumn{1}{c|}{27.9\%} & 3.5\% & \multicolumn{1}{c|}{28.5\%}        & 17.7\%        \\ \hline
Self-Experience                                                                           & o1                              & \multicolumn{1}{c|}{20.8\%} & 13.79\% & \multicolumn{1}{c|}{26.5\%}        & 17.7\%        \\ \hline
\end{tabular}
}
\end{table}

We next evaluate the impact of continuous knowledge integration (Section~\ref{sec:knowledge}) on \name's performance. Specifically, we augment \name with external documentation and execution-derived insights to dynamically improve its domain understanding without retraining. We create \textit{34} help documents tailored to benchmark tasks, each containing precise step-by-step instructions, enabling \name to retrieve the most relevant guidance (maximum of one per task) at runtime. Additionally, we implement an automated pipeline that summarizes successful execution trajectories—validated by our Task Evaluator and archives them into a retrievable knowledge database. For subsequent tasks, \name dynamically retrieves up to three relevant past execution logs to guide task planning and execution. Given that knowledge integration primarily addresses failures arising from insufficient planning (\textit{Plan Errors}), we employ the \emph{Plan Recovery Ratio (PRR)} to measure the proportion of previously failed planning cases successfully resolved by integrating new knowledge.

Table~\ref{tab:knowledge} compares the overall SR and PRRs across two benchmarks, highlighting significant performance improvements attributable to knowledge integration. Both live help-document retrieval and self-experience summarization yield noticeable gains, reducing planning failures by up to 17.7\%. Notably, self-experience enhancements using the stronger model (o1) achieve consistent improvements across both benchmarks, underscoring the efficacy of leveraging prior successes for adaptive improvement. While help documents occasionally result in modest gains, their effectiveness depends on task complexity and document specificity.

These findings underscore the value of systematic knowledge integration, demonstrating that continuous augmentation of the agent's knowledge base can substantially enhance its robustness, scalability, and adaptability in real-world deployments. Moreover, as \name continues to accumulate execution experience and documentation over time, it inherently evolves toward higher reliability and improved autonomy, marking a clear path for ongoing enhancement in desktop automation.

\subsection{Effectiveness of Speculative Multi-Action Execution} \label{sec:speculative_eval}

\begin{table}[t]
\caption{The SR and ACS comparison between single action and speculative multi-action mode.}
\label{tab:speculative}
\resizebox{\columnwidth}{!}{ 
\begin{tabular}{l|c|ccc|ccc}
\hline
\multirow{2}{*}{\textbf{\begin{tabular}[c]{@{}l@{}}Action\\ Execution\end{tabular}}} & \multirow{2}{*}{\textbf{Model}} & \multicolumn{3}{c|}{\textbf{WAA}}                                                                                                          & \multicolumn{3}{c}{\textbf{OSWorld-W}}                                                                                                     \\ \cline{3-8} 
                                                                                     &                                 & \multicolumn{1}{c|}{\textbf{SR}} & \multicolumn{1}{c|}{\textbf{ACS}}  & \textbf{\begin{tabular}[c]{@{}c@{}}Success \\ Subset\end{tabular}} & \multicolumn{1}{c|}{\textbf{SR}} & \multicolumn{1}{c|}{\textbf{ACS}}  & \textbf{\begin{tabular}[c]{@{}c@{}}Success \\ Subset\end{tabular}} \\ \hline
Single                                                                               & GPT-4o                          & \multicolumn{1}{c|}{23.4\%}      & \multicolumn{1}{c|}{10.00}         & \multirow{2}{*}{30}                                                & \multicolumn{1}{c|}{22.4\%}      & \multicolumn{1}{c|}{13.30}         & \multirow{2}{*}{10}                                                \\ \cline{1-4} \cline{6-7}
Speculative                                                                          & GPT-4o                          & \multicolumn{1}{c|}{23.4\%}      & \multicolumn{1}{c|}{\textbf{8.78}} &                                                                    & \multicolumn{1}{c|}{24.5\%}      & \multicolumn{1}{c|}{\textbf{7.40}} &                                                                    \\ \hline\hline
Single                                                                               & o1                              & \multicolumn{1}{c|}{25.3\%}      & \multicolumn{1}{c|}{9.95}          & \multirow{2}{*}{32}                                                & \multicolumn{1}{c|}{24.5\%}      & \multicolumn{1}{c|}{6.80}          & \multirow{2}{*}{10}                                                \\ \cline{1-4} \cline{6-7}
Speculative                                                                          & o1                              & \multicolumn{1}{c|}{24.7\%}      & \multicolumn{1}{c|}{\textbf{8.85}} &                                                                    & \multicolumn{1}{c|}{26.5\%}      & \multicolumn{1}{c|}{\textbf{3.30}} &                                                                    \\ \hline
\end{tabular}
}
\end{table}

\begin{figure}[t]
    \centering
    \includegraphics[width=\columnwidth]{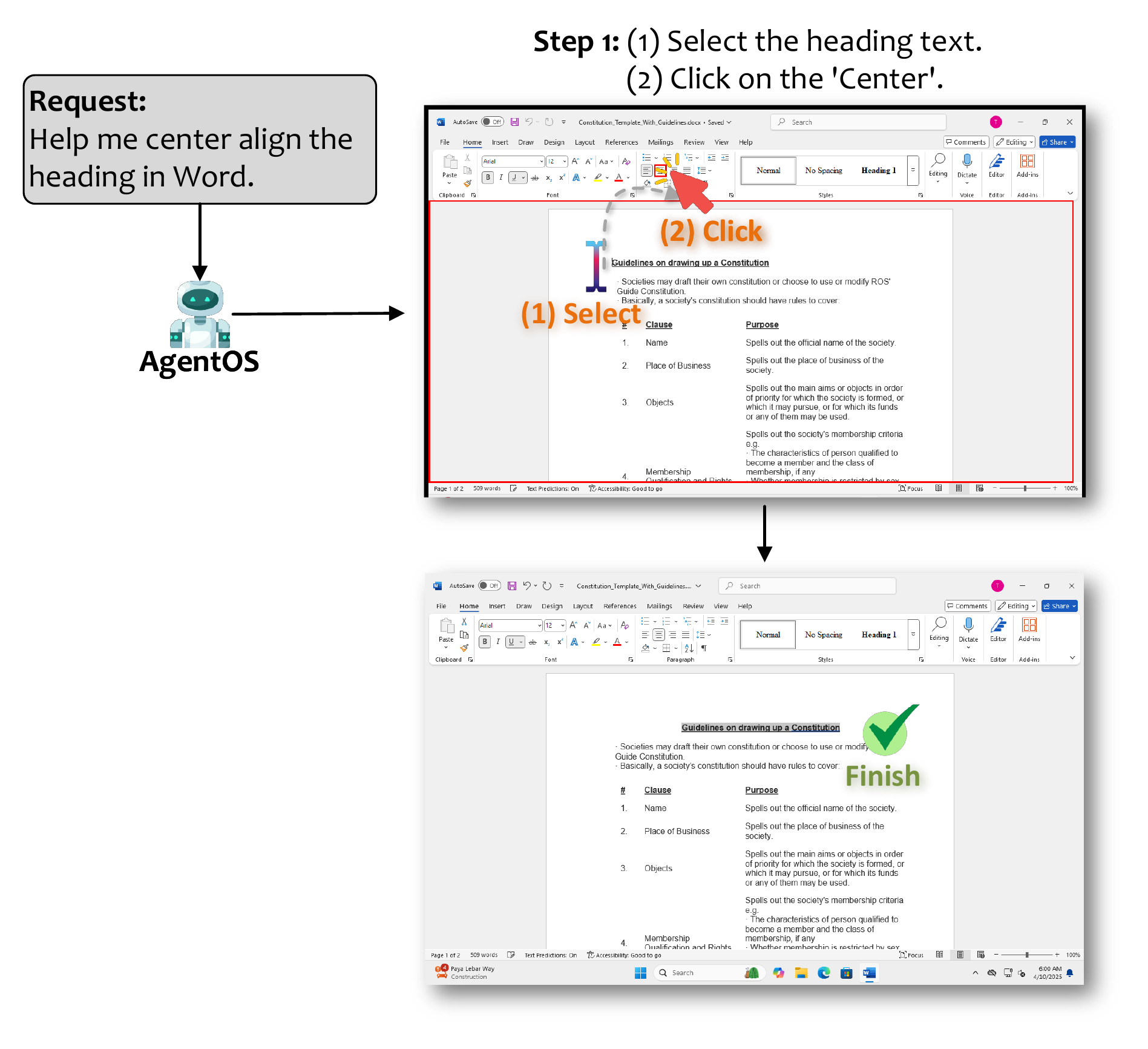}
    \vspace{-2em}
    \caption{A case study of the successful speculative multi-action execution.}
    \label{fig:multiaction_case}
\end{figure}

Next, we evaluate how speculative multi-action execution (Section~\ref{sec:speculative}) affects task completion rates and efficiency. Table~\ref{tab:speculative} compares two modes for \name: generating and executing one action per inference (\textit{single-action}) versus inferring multiple consecutive actions in one step (\textit{speculative multi-action}). To ensure a fair comparison, we compute the Average Completion Steps (ACS) only on the subset of tasks that both modes complete successfully.

The results show that speculative multi-action execution retains a comparable Success Rate (SR) to single-action mode while notably cutting the average steps—by up to 10\% on WAA and an impressive 51.5\% on OSWorld-W. Because each step requires an LLM call, reducing the number of steps significantly lowers both latency and cost. This finding confirms that speculative multi-action planning enhances efficiency without compromising reliability, further highlighting \name's ability to optimize resource utilization in practical desktop automation.

\paragraph{Case Study.} Figure~\ref{fig:multiaction_case} illustrates how speculative multi-action execution operates in practice. When the user requests that \name center-align a heading in a Word document, the sequence of steps would typically require selecting the heading text and then clicking the \texttt{Center} icon. These actions that are sequentially dependent but do not interfere with each other. Instead of treating each action as a separate LLM inference, \name predicts both actions in a single step, leveraging speculative multi-action planning. Consequently, it completes the task with just one LLM call, significantly enhancing efficiency while maintaining accuracy.


\subsection{Operator as an \appa}
\begin{figure}[t]
    \centering
    \includegraphics[width=\columnwidth]{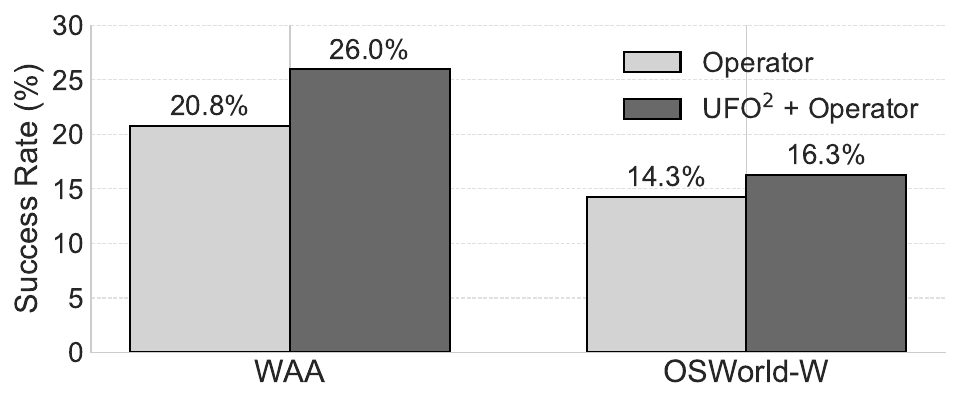}
    \vspace{-2em}
    \caption{Comparison of Operator vs. \name + Operator on WAA and OSWorld-W.}
    \label{fig:operator_comparison}
    \vspace{-1em}
\end{figure}

To demonstrate the ``Everything-as-an-\appa'' capability (Section~\ref{sec:everything}), we conducted an experiment where \name's \hosta orchestrator uses \emph{only} Operator as the \appa. In other words, all native \appas were disabled, leaving Operator to accept subtasks and communicate via the standard \name messaging protocol. The only adjustment to Operator's perception layer was restricting it to screenshots of the selected application window, rather than the full desktop.

Figure~\ref{fig:operator_comparison} shows that \name + Operator achieves higher success rates than running Operator alone, particularly on WAA (26.0\% vs. 20.8\%). We attribute these gains to three key factors. First, the \hosta breaks down complex user instructions into clearer, single-application subtasks, reducing ambiguity. Second, \hosta messages include additional tips that improve Operator's decision making. Finally, limiting Operator's view to a single, active application window reduces visual noise and simplifies control detection. Taken together, these results underscore the benefits of \name's multiagent design, while demonstrating how ``Everything-as-an-\appa'' can elevate the performance of an existing CUA.

\subsection{Efficiency Analysis}

To comprehensively understand the performance characteristics of \name, we conducted detailed profiling of task execution efficiency, focusing specifically on step count and latency.

\paragraph{Step Count Profiling.}
\begin{table*}[t]
\caption{Step count statistic for \name.}
\label{tab:step}
\begin{tabular}{l|l|cccc|cccc}
\hline
\multirow{2}{*}{\textbf{Agent}} & \multirow{2}{*}{\textbf{Model}} & \multicolumn{4}{c|}{\textbf{WAA}}                                                                                                                                        & \multicolumn{4}{c}{\textbf{OSWorld-W}}                                                                                                                                   \\ \cline{3-10} 
                                &                                 & \multicolumn{1}{c|}{\hosta}        & \multicolumn{1}{c|}{\appa}         & \multicolumn{1}{c|}{Total}          & \begin{tabular}[c]{@{}c@{}}Success\\ Subset\end{tabular} & \multicolumn{1}{c|}{\hosta}        & \multicolumn{1}{c|}{\appa}         & \multicolumn{1}{c|}{Total}          & \begin{tabular}[c]{@{}c@{}}Success\\ Subset\end{tabular} \\ \hline
\name-base                      & GPT-4o                          & \multicolumn{1}{c|}{\textbf{2.21}} & \multicolumn{1}{c|}{8.11}          & \multicolumn{1}{c|}{10.32}          & \multirow{2}{*}{31}                                      & \multicolumn{1}{c|}{\textbf{1.80}} & \multicolumn{1}{c|}{10.80}         & \multicolumn{1}{c|}{12.60}          & \multirow{2}{*}{7}                                       \\ \cline{1-5} \cline{7-9}
\name                           & GPT-4o                          & \multicolumn{1}{c|}{2.32}          & \multicolumn{1}{c|}{\textbf{7.89}} & \multicolumn{1}{c|}{\textbf{10.21}} &                                                          & \multicolumn{1}{c|}{2.80}          & \multicolumn{1}{c|}{\textbf{7.20}} & \multicolumn{1}{c|}{\textbf{10.00}} &                                                          \\ \hline\hline
\name-base                      & o1                              & \multicolumn{1}{c|}{2.14}          & \multicolumn{1}{c|}{7.00}          & \multicolumn{1}{c|}{9.14}           & \multirow{2}{*}{34}                                      & \multicolumn{1}{c|}{2.50}          & \multicolumn{1}{c|}{8.83}          & \multicolumn{1}{c|}{11.33}          & \multirow{2}{*}{8}                                       \\ \cline{1-5} \cline{7-9}
\name                           & o1                              & \multicolumn{1}{c|}{\textbf{2.00}} & \multicolumn{1}{c|}{\textbf{4.05}} & \multicolumn{1}{c|}{\textbf{6.05}}  &                                                          & \multicolumn{1}{c|}{\textbf{2.00}} & \multicolumn{1}{c|}{\textbf{3.50}} & \multicolumn{1}{c|}{\textbf{5.50}}  &                                                          \\ \hline
\end{tabular}
\end{table*}

Table~\ref{tab:step} summarizes the average number of execution steps performed by the \hosta and \appas across both benchmark suites. The steps reported are computed on tasks that were successfully completed across all model configurations, ensuring fair comparisons. Two key insights emerge:

First, the fully integrated \name configuration consistently reduces the average number of steps required compared to the baseline (\name-base), achieving reductions of up to 50\%. This substantial efficiency gain demonstrates how deep OS integration, specifically the hybrid GUI–API action orchestration and advanced control detection strategies, significantly streamline execution paths.

Second, utilizing a more powerful reasoning model (\eg, \textit{o1} versus GPT-4o) further reduces step counts, indicating that enhanced reasoning capability enables the agent to identify and exploit more efficient action sequences. For instance, stronger models can better leverage direct API interactions or avoid unnecessary intermediate GUI interactions. This underscores the complementary role of both robust system integration and advanced LLM reasoning in minimizing execution overhead.

\paragraph{Latency Breakdown.}
\begin{figure}[t]
    \centering
    \includegraphics[width=\columnwidth]{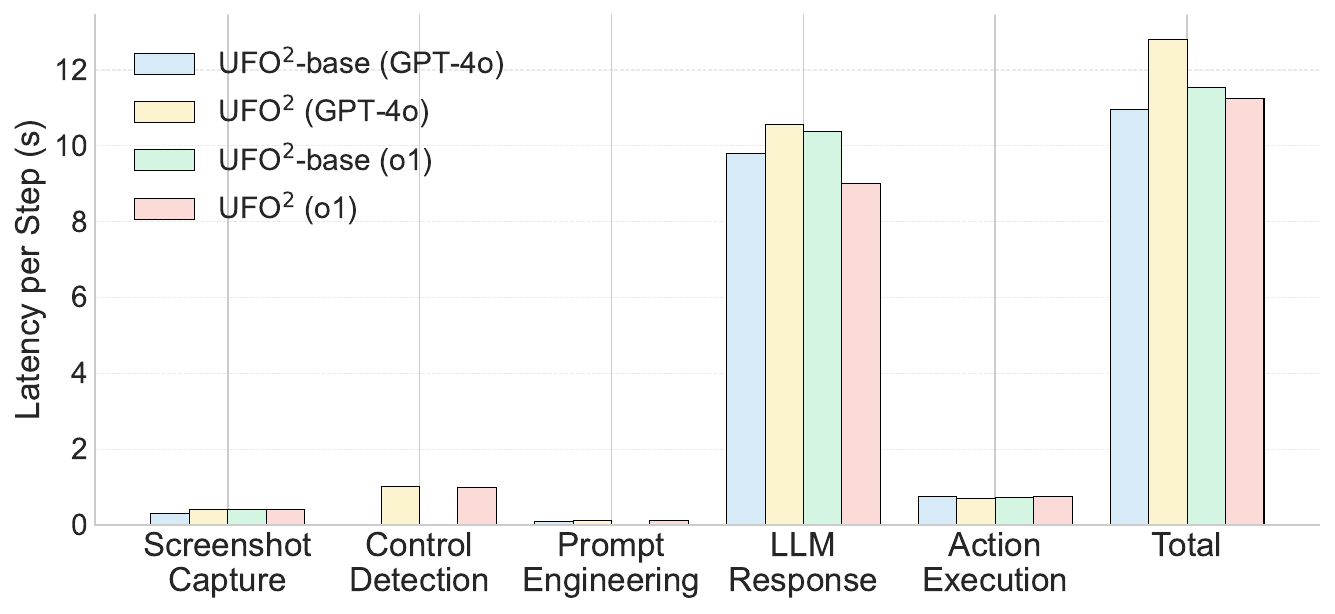}
    \vspace{-2em}
    \caption{Average time cost per-stage of a single execution step.}
    \label{fig:latency}
    \vspace{-1em}
\end{figure}

Figure~\ref{fig:latency} provides a detailed breakdown of average latency per execution step in \name, separated into five key phases:
\textit{(i)} \emph{Screenshot Capture},  
\textit{(ii)} \emph{Control Detection} via UIA APIs (and) OmniParser-v2,  
\textit{(iii)} \emph{Prompt Preparation}, including retrieval of relevant help documents and historical execution experiences,  
\textit{(iv)} \emph{LLM Inference}, and  
\textit{(v)} \emph{Action Execution} on target applications.

Across all configurations, the LLM inference phase dominates total latency, averaging around 10 seconds per inference. This bottleneck highlights a clear opportunity for optimization by deploying smaller, specialized models or employing more powerful inference hardware—strategies that remain viable but are beyond our current evaluation scope.

Excluding LLM inference overhead, the baseline system (\name-base) achieves highly efficient execution, incurring around 10 seconds per step on average. In contrast, the fully integrated \name incurs only an additional 1 second per step for its hybrid control detection pipeline, largely due to OmniParser-v2 visual parsing. This added overhead represents a deliberate trade-off, significantly enhancing the robustness and accuracy of GUI control detection at a modest latency cost.

Taken together, these results indicate that the substantial reduction in total steps required per task ensures overall task completion times remain practical (approximately 1 minute per task). These profiling insights reinforce that \name's comprehensive system-level integration balances latency, accuracy, and efficiency, offering a scalable and performant solution for real-world desktop automation.

\subsection{Model Ablation}


\begin{figure}[t]
    \centering
    \includegraphics[width=\columnwidth]{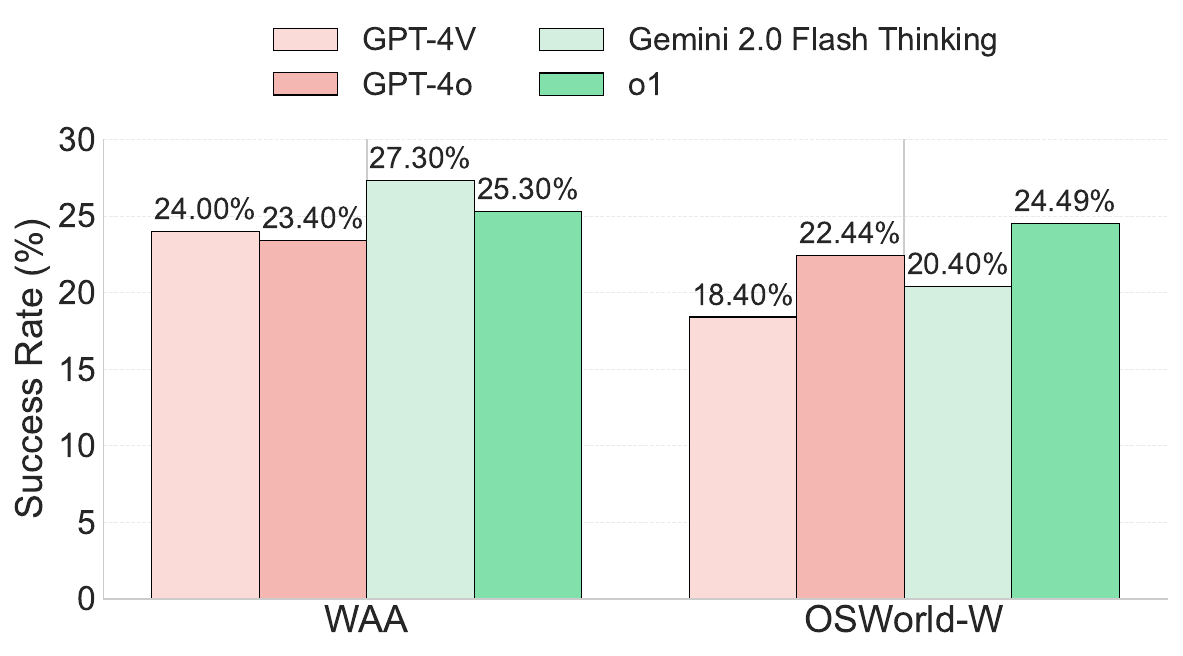}
    \vspace{-2em}
    \caption{Comparison of different LLMs used in \name and \name-base on WAA and OSWorld-W.}
    \label{fig:model}
    \vspace{-1em}
\end{figure}

Table~\ref{fig:model} compares the performance of \name and \name-base across four large language models. GPT-4V and GPT-4o generate direct answers without exposing an explicit CoT, whereas Gemini 2.0 (Flash Thinking) and o1 embed reasoning steps internally before producing final outputs. Overall, models with built-in reasoning typically achieve higher success rates (SR), highlighting the value of more deliberative or CoT-driven processes in desktop automation~\cite{lu2025uir1enhancingactionprediction}.

This result underscores a promising direction for CUAs: fine-tuning advanced reasoning models specifically for desktop automation tasks. By allowing agents to formulate and refine multi-step plans—especially when integrated with deeper OS-level signals—\name can address complex or ambiguous situations more reliably. As LLM-based reasoning continues to mature, we expect further gains in both accuracy and generality from \name's model-agnostic design.



\section{Discussion \& Future Work}

\paragraph{Latency and Responsiveness.}
\name currently invokes LLM inference at each decision step, incurring a latency typically ranging from several seconds to tens of seconds per action. Despite various engineering optimizations, complex tasks comprising multiple sequential actions can accumulate to execution times of 1--2 minutes, which remains acceptable but still inferior to skilled human performance. To alleviate user-perceived delay, we introduced the Picture-in-Picture (PiP) interface, enabling \name to execute tasks unobtrusively within an isolated virtual desktop, thus substantially reducing user inconvenience during longer-running automations. In future work, we aim to further lower latency by investigating the deployment of specialized, lightweight Large Action Models (LAMs)~\cite{wang2024large}, optimized for task-specific inference to enhance both responsiveness and scalability.

\paragraph{Closing the Gap with Human-Level Performance.}
Our comprehensive evaluations indicate that \name, while robust and effective, does not yet consistently achieve human-level performance across all Windows applications. Bridging this gap will necessitate advances primarily along two critical dimensions. First, enhancing foundational visual-language models through fine-tuning on extensive, diverse GUI interaction datasets will significantly improve agents' capabilities and generalization across varied applications. Second, tighter integration with OS-level APIs, native application interfaces, and comprehensive, structured documentation sources will deepen contextual understanding and bolster execution reliability. Given \name's modular architecture, these enhancements can be incrementally adopted, continuously refining performance towards human-equivalent proficiency across diverse application scenarios.

\paragraph{Generalization Across Operating Systems.}
While \name targets Windows OS due to its widespread market adoption (over 70\% market share\footnote{\url{https://gs.statcounter.com/os-market-share/desktop/worldwide/\#monthly-202301-202301-bar}}), the modular, layered system design facilitates straightforward adaptation to other desktop platforms, such as Linux and macOS. The core approach—leveraging accessibility frameworks like Windows UI Automation (UIA)—has direct analogs on Linux (AT-SPI~\cite{atspi}) and macOS (Accessibility API~\cite{apple_accessibility_api}). Consequently, the existing design principles, agent decomposition strategy, and unified GUI-API orchestration model generalize naturally, enabling rapid, platform-specific customizations. Exploring cross-platform deployments will be an important area of future work, potentially laying the groundwork for a unified ecosystem of desktop automation solutions spanning diverse operating environments.

\section{Related Work}

Integrating LLMs into OS represents a growing, yet nascent area of research. In this section, we discuss prior work that intersects with our research on system-level integration of multimodal LLM-based desktop automation agents.

\subsection{Computer-Using Agents (CUAs)}

Recent advancements in multimodal LLMs have significantly accelerated the development of \emph{Computer-Using Agents} (CUAs), which automate desktop workflows by simulating GUI interactions at the OS level. Early pioneering systems, such as UFO~\cite{zhang2024ufo}, leveraged multimodal models (\eg GPT-4V~\cite{yang2023dawn}) alongside UIA APIs to interpret graphical interfaces and execute complex tasks via natural language instructions. UFO notably introduced a multi-agent architecture, enhancing the reliability and capability of CUAs to handle cross-application and long-term workflows.

Subsequent efforts have focused primarily on refining underlying multimodal models and extending platform capabilities. For example, CogAgent~\cite{hong2024cogagent}, built upon the vision-language model CogVLM~\cite{wang2024cogvlm}, specialized in GUI understanding across multiple platforms (PC, web, Android), representing one of the earliest dedicated multimodal CUAs. Industry interest has similarly accelerated with Anthropic's Claude-3.5 (Computer Use)~\cite{anthropic2024}, an agent relying entirely on screenshot-based GUI interactions, and OpenAI's Operator~\cite{cua2025}, which significantly improved desktop automation performance through advanced multimodal reasoning.

However, these existing CUAs remain largely prototype demonstrations, often lacking deep integration with the OS and native application capabilities. In contrast, our work in \name directly addresses these fundamental system-level limitations through a modular AgentOS architecture, deep OS and API integration, hybrid GUI detection, and a non-disruptive execution model, bridging the gap between conceptual CUAs and practical desktop automation.

\subsection{LLMs for Operating Systems}

Another promising research direction involves embedding LLMs directly within OS architectures, aiming to substantially enhance automation, adaptability, and usability. Ge \etal first proposed the conceptual framework AIOS~\cite{ge2023llm}, positioning an LLM at the center of OS design to orchestrate high-level user interactions and automated decision-making. In their vision, agents resemble OS applications, each exposing specialized capabilities accessible via natural language, effectively enabling users to "program" their OS intuitively.

Building on this conceptual foundation, Mei \etal~\cite{mei2024aios} realized AIOS as a concrete prototype, encapsulating LLM interactions and tool APIs within a privileged OS kernel. This design provides core OS functionalities such as process scheduling, memory management, I/O handling, and access control, leveraging LLMs to simplify agent development through a dedicated SDK. Rama \etal~\cite{rama2025cerebrumaiossdkplatform} extended this paradigm, introducing semantic file management capabilities directly within traditional OS environments through AIOS-based agents, further demonstrating practical system-level integration.

Complementing these high-level OS integrations, AutoOS~\cite{chen2024autoos} applied LLMs for automatic tuning of kernel-level parameters in Linux, achieving substantial efficiency gains through autonomous exploration and optimization. This highlights another dimension where LLM integration can directly enhance core system performance and management.

Collectively, these research efforts illustrate an emerging paradigm shift where LLMs become integral components of operating systems, enabling powerful automation, enhanced user interaction, and adaptive system behavior. Our work with \name extends this line of research specifically to desktop automation, offering a deeply integrated, scalable, and practical AgentOS that leverages multimodal LLMs in conjunction with robust OS-level mechanisms.

\section{Conclusion}
We introduced \textbf{\name}, a practical, OS-integrated Windows desktop automation AgentOS that transforms CUAs from conceptual prototypes into robust, user-oriented solutions. Unlike prior CUAs, \name leverages deep system-level integration through a modular, multiagent architecture consisting of a centralized \hosta and application-specialized \appas. Each \appa seamlessly combines GUI interactions with native APIs and continually integrates application-specific knowledge, substantially improving reliability and execution efficiency. \name can operate on a PiP virtual desktop interface further enhances usability, enabling concurrent user-agent workflows without interference.

Our comprehensive evaluation across over 20 real-world Windows applications demonstrated that \name achieves significant improvements in robustness, accuracy, and scalability compared to state-of-the-art CUAs. Notably, by coupling our integrated framework with robust OS-level features, even less specialized foundation models (\eg GPT-4o) surpass specialized CUAs such as Operator.

\bibliographystyle{unsrt}
\bibliography{sample-base}



\end{document}